\documentclass[10pt,twocolumn,letterpaper]{article}
\usepackage{color} 
\usepackage[accsupp]{axessibility} 

\makeatletter
\DeclareRobustCommand\onedot{\futurelet\@let@token\@onedot}
\def\@onedot{\ifx\@let@token.\else.\null\fi\xspace}
\def\eg{\emph{e.g}\onedot} 
\def\ie{\emph{i.e}\onedot} 
 
\def\etc{\emph{etc}\onedot} 
 
\def\etal{\emph{et al}\onedot}
\makeatother
\newcommand{\Tref}[1]{Table~\ref{#1}}

\newcommand{\Fref}[1]{Figure~\ref{#1}}

\newcommand{\sref}[1]{Sec.~\ref{#1}}

\usepackage{cvpr}              %
\usepackage[dvipsnames]{xcolor}
\usepackage{cuted}
\definecolor{cvprblue}{rgb}{0.21,0.49,0.74}
\usepackage{transparent}
\usepackage[pagebackref,breaklinks,colorlinks,citecolor=cvprblue]{hyperref}
\usepackage{multirow}
\usepackage[capitalize]{cleveref}
\crefname{section}{Sec.}{Secs.}
\Crefname{section}{Section}{Sections}
\Crefname{table}{Table}{Tables}
\crefname{table}{Tab.}{Tabs.}

\makeatletter
\renewcommand{\paragraph}{%
  \@startsection{paragraph}{4}%
  {\z@}{0.2ex \@plus 0.3ex \@minus .2ex}{-1em}%
  {\normalfont\normalsize\bfseries}%
}
\makeatother

\def\tbfta{7pt}
\def\tbftb{8pt} 

\def\paperID{138} %
\def\confName{CVPR}
\def\confYear{2024}

\title{Real-time 3D-aware Portrait Video Relighting}

\author{Ziqi Cai\textsuperscript{1,2}~~~
Kaiwen Jiang\textsuperscript{3}~~~
Shu-Yu Chen\textsuperscript{1}~~~
Yu-Kun Lai\textsuperscript{4}~~~
Hongbo Fu\textsuperscript{5,6}~~~
Boxin Shi\textsuperscript{8,9}~~~
Lin Gao\thanks{Corresponding author is Lin Gao}~\textsuperscript{1,7}
\vspace{2mm}
\\
\parbox{\textwidth}{\centering \small
{\textsuperscript{1}Beijing Key Laboratory of Mobile Computing and Pervasive Device, Institute of Computing Technology, Chinese Academy of Sciences}\\
{\textsuperscript{2}Beijing Jiaotong University}\quad
{\textsuperscript{3}University of California San Diego}\quad
{
{\textsuperscript{4}Cardiff University}\quad
{\textsuperscript{5}City University of Hong Kong}\\
{\textsuperscript{6}The Hong Kong University of Science and Technology}\quad
{\textsuperscript{7}University of Chinese Academy of Sciences}\\
{\textsuperscript{8}National Key Laboratory for Multimedia Information Processing, School of Computer Science, Peking University}\\
{\textsuperscript{9}
National Engineering Research Center of Visual Technology, School of Computer Science, Peking University}}
\vspace{2mm}
\\
{\tt\small \{zqtsai,kevinjiangedu\}@gmail.com},~~
{\tt\small chenshuyu@ict.ac.cn},~~
{\tt\small Yukun.Lai@cs.cardiff.ac.uk}\vspace{-1mm}
\\
{\tt\small hongbofu@cityu.edu.hk},~~
{\tt\small shiboxin@pku.edu.cn},~~
{\tt\small gaolin@ict.ac.cn}
}
}

\begin{document}

\maketitle

\begin{abstract}
Synthesizing realistic videos of talking faces under custom lighting conditions and viewing angles benefits various downstream applications like video conferencing. 
However, most existing relighting methods are either time-consuming or unable to adjust the viewpoints.
In this paper, we present the first real-time 3D-aware method for relighting in-the-wild videos of talking faces based on Neural Radiance Fields (NeRF). 
Given an input portrait video, our method can synthesize talking faces under both novel views and novel lighting conditions with a photo-realistic and disentangled 3D representation.
Specifically, we infer an albedo tri-plane, as well as a shading tri-plane 
based on a desired lighting condition 
for each video frame with fast dual-encoders. %
We also leverage a temporal consistency network to ensure smooth transitions and reduce flickering artifacts. 
Our method runs at 32.98 fps on consumer-level hardware and achieves state-of-the-art results in terms of reconstruction quality, lighting error%
, lighting instability, temporal consistency and inference speed. We demonstrate the effectiveness and interactivity of our method on various portrait videos with diverse lighting and viewing conditions.
\end{abstract}

\section{Introduction}
\label{sec:Introduction}
Portrait videos are widely used in various scenarios, such as video conferencing, video editing, entertainment, virtual reality, \etc.
However, many portrait videos are captured under unsatisfactory conditions, such as environments that are either too dark or too bright, or with virtual backgrounds that do not match the lighting of the foreground. These factors degrade the visual quality and realism of videos and affect the user experience.

\begin{figure}[t]
\centering
\includegraphics[width=\linewidth]{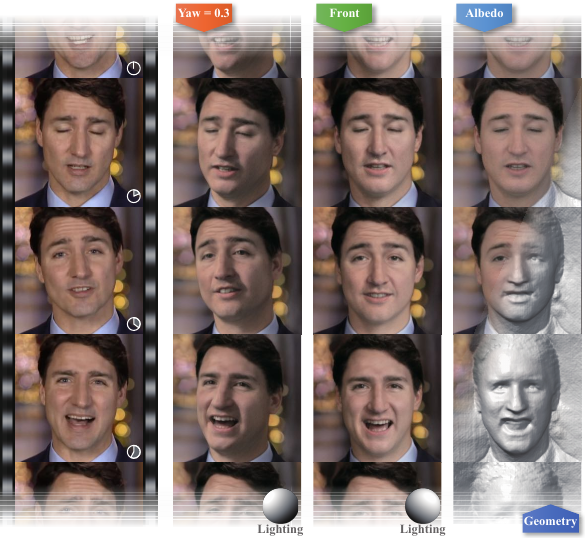} 
\captionof{figure}{
Given a portrait video shown in the leftmost column, our method reconstructs a 3D relightable face for each video frame. Users can then adjust {their}
viewpoints and lighting conditions interactively. 
{The second column displays relighted video frames with a head pose yaw of 0.3%
, while the third column presents faces relighted under an alternative lighting condition with a frontal head pose. The rightmost column provides the predicted albedo and geometry of the reconstructed face.}
Please see the supplementary video for the full results.
} \label{fig:teaser}
\end{figure}

Of particular significance is the context of augmented reality (AR) and virtual reality (VR) applications, where users often seek to create 3D faces that can be dynamically relighted to fit %
the environment. This dynamic relighting capability becomes possible only when the underlying method is inherently 3D-aware and operates in real time.

However, 3D-aware portrait video relighting is a challenging task, since it involves modeling the complex interactions between the light, geometry, and appearance of human faces, as well as ensuring the temporal coherence and naturalness of synthesized videos.
It is even more challenging when real-time performance is required. Existing methods for face relighting suffer from some limitations that prevent them from being widely adopted in practice. 
First, most of them (\eg, \cite{yeh2022learning,pandey2021total,Zhang_2021_ICCV}) can only relight the faces from the input viewpoints, thus restricting %
the user’s freedom to change the camera angle or perspective. This also limits the creative possibilities and applications for AR/VR scenarios. 
Second, many methods (\eg, \cite{DPR,SMFR}) are designed for monocular {image} inputs %
and thus produce flickering or unnatural results when directly applied to videos, making them inferior for practical usage, where smooth and realistic transitions are expected. 
Third, some methods are time-consuming in terms of both training and inference.
For example, ReliTalk~\cite{qiu2023relitalk} takes 3 days of training for a %
{2-minute} video clip. {Once trained, it takes 0.2 seconds to relight a video frame.} %
Although DPR~\cite{DPR} achieves real-time performance, it suffers from low-quality results. It is still challenging to balance quality and efficiency with existing solutions.

In this paper, we present %
{a novel real-time 3D-aware portrait video relighting method that jointly solves the above problems}
by generating realistic and consistent relighting results for faces from novel viewpoints in real time, enabling users to create realistic and natural personas for AR/VR applications{, as shown in \Fref{fig:teaser}.} %
In summary, our {technical} contributions are:
\begin{itemize} 
\item We contribute to the ongoing field of 3D-aware portrait video relighting by introducing a novel approach that achieves real-time performance while producing realistic and consistent results.

\item We propose to use dual feed-forward encoders to capture the albedo and shading information within a portrait. The shading encoder is conditioned on the albedo encoder to ensure spatial alignment of albedo and shading, resulting in realistic reconstruction and accurate relighting.

\item We use a novel temporal consistency network to address temporal inconsistencies in video data, reducing flickering artifacts and ensuring seamless transitions between frames.
\end{itemize}

\section{Related Work}
\label{sec:Related Work}

\begin{figure*}[t]
    \centering
\includegraphics[width=1\linewidth]{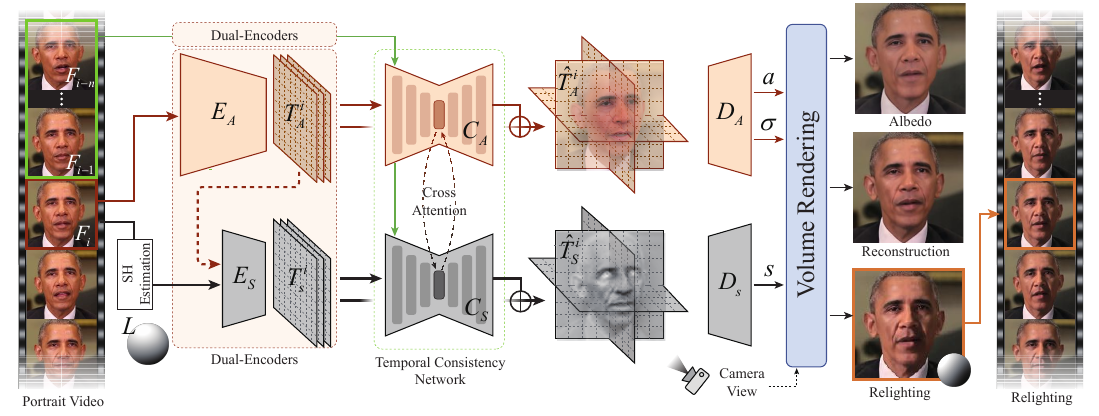}
\caption{The pipeline of our method. Given a portrait video {shown on the left side}, we embed each video frame into {an albedo tri-plane and a shading tri-plane} 
using {Dual-Encoders}.
For example, for frame {$F_i$}, we predict the {albedo tri-plane $T_A^i$.} %
Next, we use the estimated lighting condition $L$ and the {albedo} %
{tri-plane $T_A^i$} to predict the {shading tri-plane $T_S^i$} %
that {models} %
the illumination effects on the face. Then we feed $T_S^i$ and $T_A^i$ 
along with the tri-planes predicted from previous $n$ frames %
into two transformer models {$C_A$ and $C_S$} %
to enhance the temporal consistency. The two transformers use cross-attention to cooperate for information sharing and alignment between the albedo and shading branches. We add the predicted residual %
to {$T_A^i$} and {$T_S^i$} {as $\hat{T}_A^i, \hat{T}_S^i$} for better temporal consistency. Finally, we use {$\hat{T}_A^i$} and {$\hat{T}_S^i$} to condition the volumetric rendering process, producing depth, albedo, shading, color, and super-resolved images.}%
    \label{fig:pipeline}
\end{figure*}

Our work closely relates to several topics, including 3D-aware portrait generation, portrait relighting, and GAN (Generative Adversarial Network) inversion.

\subsection{3D-aware Portrait Generation}

3D-aware portrait generation is the task of generating realistic and diverse images of human faces. Previous work on this task relied on 3D face priors to model the geometry and appearance of faces, such as 3D morphable models~\cite{blanz:1999:morphable,FLAME:SiggraphAsia2017} or neural face models~\cite{tewari2017mofa,DECA:Siggraph2021}. However, these methods require expensive 3D scanning or manual annotation and often produce low-resolution or unnatural results. With the advancement of generative models~\cite{sun2024recent}, it is now possible to learn a 3D representation of faces from a collection of 2D images without any explicit 3D supervision. In particular, recent approaches combine neural radiance fields (NeRF)~\cite{nerf} and generative models such as generative adversarial networks (GANs)~\cite{goodfellow2020generative} and diffusion models~\cite{ho2020denoising} to generate high-resolution and multi-view consistent face images~\cite{chan2022efficient,panohead_2023_CVPR,Schwarz2020NEURIPS,Niemeyer2020GIRAFFE,pan2021shadegan,deng2022gram,xue2022giraffehd,sun2022ide,muller2023diffrf,sun2023next3d,wang2023rodin}. In this paper, we adopt the tri-plane representation from EG3D~\cite{chan2022efficient} as our 3D representation for portrait synthesis and relighting.
{This choice is motivated by the insights presented in \cite{nerffacelighting}, which shows 
that the tri-plane 3D representation facilitates the disentanglement of albedo and shading. This disentanglement, afforded by the tri-plane structure, enables a 3D-aware approach for relighting portraits in a photorealistic manner.}

\subsection{Portrait Relighting}\label{subsec:Portrait Relighting}

Portrait relighting requires changing the illumination of a portrait image or video while preserving the identity and appearance of the subject. Previous works ({\eg},~\cite{Zhang_2021_ICCV}) used One-Light-at-A-Time (OLAT) capturing systems to obtain detailed portrait geometry and reflectance, which enabled realistic relighting results. However, OLAT data is expensive and difficult to acquire, thus limiting %
the applicability of these methods. To overcome this limitation, some recent works (\eg,~\cite{zhou2019deep,yeh2022learning,papantoniou2023relightify,fei2023split}) used synthetic data for training and showed good generalization to real data.

Another line of research explored 3D-aware portrait relighting, which leveraged the recent advances in unconditional 3D-aware portrait generation~\cite{chan2022efficient} by combining %
{GANs}~\cite{goodfellow2020generative} and {NeRFs}~\cite{nerf}. Concurrently, Jiang \etal~\cite{nerffacelighting} and Ranjan \etal~\cite{ranjan2023} modeled the lighting effects in generative models, either implicitly or explicitly, and achieved impressive quality of image relighting. However, these methods are unsuitable for video relighting since they require inverting each frame separately, which is time-consuming and does not ensure temporal consistency, leading to flickering artifacts.

This paper proposes a novel method for real-time 3D-aware video relighting, which builds on~\cite{nerffacelighting} and distills its knowledge into a feedforward network with a temporal enhancement module. Our method can produce realistic, high-quality portrait relighting videos with various lighting effects and novel views. In contrast to our approach, none of the existing portrait relighting techniques can handle consistent and real-time novel view synthesis for a video sequence.%

\subsection{GAN Inversion}\label{subsec:GAN Inversion}
GAN inversion aims to find a latent representation in a pretrained model's latent space that can reconstruct a given image with its generator. Existing GAN inversion methods can be divided into {optimization-based, learning-based, and hybrid approaches.}

{Optimization-based methods minimize reconstruction errors for high-quality results but are slow, as seen in \cite{yin2022spi} and \cite{xie2022high}, since these methods require end-to-end optimization across numerous video frames. %
Learning-based methods (\eg, \cite{tov2021designing}), using an encoder, are faster but at the cost of %
lower-quality reconstruction quality. %
With recent trends of predicting richer information from input images, Yuan \etal~\cite{yuan2023make}, Bhattarai \etal~\cite{bhattarai2023triplanenet}, and Trevithick \etal~\cite{LP3D} propose to predict a tri-plane from input images, striking a good balance between quality and efficiency. Hybrid methods (\eg, \cite{roich2022pivotal,yin2022nerfinvertor,bai2022high,Fruehstueck2023VIVE3D}) combine optimization and learning, enhancing both quality and efficiency. Nevertheless, their practical utility is hindered because they still require minutes to hours to process a video clip, preventing real-time applications.}

Among these methods, only pure learning-based methods have the potential for real-time applications. Based on the idea of LP3D~\cite{LP3D}, we propose a novel learning-based method for video inversion, which predicts tri-plane representations from input images instead of latent codes. Tri-plane representations contain richer information than latent codes and can better capture the geometry and appearance variations of the input images. Unlike previous learning-based methods{, such as \cite{tov2021designing,yuan2023make,bhattarai2023triplanenet,LP3D},} that are designed for single-image inversion and thus neglect the temporal information in videos, we introduce a temporal consistency network to enforce smooth transitions between consecutive frames. Our method can achieve high-quality and consistent video inversion in real time with relighting capabilities.

\section{Methodology}
\label{sec:methodology}
In this section, we give the preliminaries of the pre-trained generator in \sref{subsec:Preliminaries}. Then, we describe how we achieve real-time video inversion and enable lighting control by using two tri-planes in \sref{subsec:Tri-plane Encoder}. Next, we introduce how to enhance temporal consistency for video inputs in \sref{subsec:Temporal Consistent Network}. 
Finally, we introduce our training objectives in \sref{subsec:Training Objectives}. The overall pipeline is illustrated in \Fref{fig:pipeline}.

\subsection{Preliminaries}
\label{subsec:Preliminaries}

Our work distills knowledge from a pre-trained 3D-aware generator $G$ %
trained based on the GAN framework~\cite{nerffacelighting}, to enable real-time synthesis and lighting control of multiview consistent video frames. 
Given a latent code $w$ in an albedo %
latent space, {an albedo} %
tri-plane is first {predicted through a generator} and %
{then fed into a convolutional network \cite{stylegan2} to predict} a shading tri-plane{, which is additionally conditioned on the }%
second-order spherical harmonic (SH) coefficients {$L$} \cite{10.1145/383259.383317}. Both {albedo} %
tri-plane and shading tri-plane are used to condition the neural rendering process {given a viewing angle}. In this way, a realistic facial image $I$ and its corresponding albedo $A$ can be generated, while allowing the disentangled control of camera
and lighting conditions.

\subsection{Tri-plane {Dual-encoders}}
\label{subsec:Tri-plane Encoder}

We present dual-encoders (\Fref{fig:pipeline}) that can infer an albedo tri-plane and a shading tri-plane from a single RGB image. These two tri-planes are later rendered into a high-resolution ($512 \times 512$) RGB image $\hat{I}$ and an albedo image $\hat{A}$ through a rendering process identical to \cite{nerffacelighting}. Our network extends the LP3D model~\cite{LP3D}, which encodes an image into a tri-plane representation for neural rendering. However, unlike LP3D, our network can produce two disentangled tri-planes, allowing for dynamic adjustments of lighting conditions %
from a single image. Our network consists of two branches: one {is Albedo Encoder $E_A$} for inferring {an albedo} %
tri-plane that captures the shape and texture of the scene, and the other {is Shading Encoder $E_S$}  for inferring a shading tri-plane that models the fine-grained illumination effects.

\paragraph{Albedo Encoder.}
Inspired by LP3D~\cite{LP3D}, we use an encoder based on Vision Transformer (ViT)~\cite{dosovitskiy2020vit} in the {albedo} %
branch for albedo prediction. The input to our method is a single RGB image {$F$} %
with an overlaid coordinate map, forming a 5-channel image. We use a DeepLabV3~\cite{chen2017rethinking} network pretrained on ImageNet~\cite{deng2009imagenet} to extract low-frequency features from the input image, which capture global context and semantic information. We then feed these features into a ViT-based encoder \cite{LP3D} that further enhances the global features by self-attention mechanisms to get final low-frequency feature {$f_{\text{low}}$}. 
We also use a convolutional neural network (CNN)~\cite{LP3D} to extract high-frequency features {$f_{\text{high}}$} from the input image {$F$}, which capture the fine details and edges. We feed {$f_{\text{high}}$} into another ViT-based encoder \cite{LP3D}, along with the low-frequency features {$f_{\text{low}}$} %
to predict the final {albedo} %
tri-plane $T_{\text{A}}$. %

\paragraph{Shading Encoder.}
To predict the shading tri-plane $T_{S}$, we use a CNN with additional StyleGAN~\cite{stylegan2} blocks, conditioned on the {albedo} %
tri-plane $T_{A}$ and the lighting condition $L$. We represent the lighting condition $L$ as second-order %
SH coefficients mapped using an off-the-shelf mapping network~\cite{nerffacelighting}. This design ensures that the shading tri-plane %
$T_{S}$ is spatially aligned with the {albedo} %
tri-plane %
$T_{A}$ %
for realistic reconstruction and relighting.

We employ a {three}-stage training strategy for our encoder. {In the initial stage, we adhere to the procedure outlined in \cite{LP3D} to train the albedo encoder, focusing on reconstructing the provided portrait without considering the disentanglement between albedo and shading.} In the second stage, we independently train the albedo branch and the shading branch. In the third stage, we integrate the two branches and train them jointly. This strategic approach enhances convergence and performance compared to training both branches simultaneously from the outset.

\subsection{Temporal Consistency Network}
\label{subsec:Temporal Consistent Network}

We aim to invert a video sequence into a sequence of tri-planes, which are low-dimensional representations of the 3D scene structure, texture, and illumination. However, simply inverting each video frame independently leads to temporal inconsistency and causes flickering artifacts in the rendered images. 

To address this problem, we propose a temporal consistency network {(\Fref{fig:pipeline})}, which exploits the rich temporal information in the video sequence to enhance the temporal consistency of the tri-plane features. The network is composed of two transformers, denoted as $C_A$ and $C_S$, accompanied by an additional convolutional neural network (CNN). Our design is inspired by \cite{Lai-ECCV-2018}, yet distinctively employs features at the tri-plane level.
{Both transformers take in corresponding predicted tri-planes for $n$ frames, and predict residual tri-planes for each frame $i$ to be added to the original tri-planes as $\hat{T}^i_A, \hat{T}^i_S$.} 
The residual tri-planes capture the temporal variations and dynamics of the subject and help to eliminate the flickering effects. Moreover, {this network} %
uses cross-attention between the {albedo} %
branch and the shading branch, which allows them to interact with each other 
for better temporal consistency.

We use synthetic data to train such a temporal consistency network. Similar to training the tri-plane encoder, we generate synthetic data with augmentation techniques tailored for temporal consistency. This involves interpolating between two randomly selected camera views to simulate realistic video sequences. Additionally, random noise is added to both tri-planes to emulate flickering effects. This process for generating synthetic data provides us with a ground truth for de-flickering, {devoid of errors stemming from inaccurate camera and lighting estimations.} 
{We empirically find that such a temporal consistency network trained on dynamic viewing angles and artificial noises make our method robust towards more diverse temporal dynamics in the real-world case, such as dynamic expressions.}

\subsection{Training Objectives}
\label{subsec:Training Objectives}
\setlength{\abovedisplayskip}{0pt}
\setlength{\belowdisplayskip}{0pt}
{We first train our tri-plane dual-encoders to converge, and then train the temporal consistency network. }
{Specifically, the tri-plane dual-encoders are trained with loss defined as follows:}
\paragraph{Albedo Loss.} This loss quantifies the dissimilarity between the predicted and ground-truth albedo images and tri-planes. %
Specifically, the albedo loss is defined as:
\begin{equation}
\label{equation:albedo loss}
\begin{aligned}
\mathcal{L}_\text{albedo} &= ||\hat{A}-A||_{1}
+ ||\hat{A_{r}}-A_{r}||_{1}
+ \mathcal{L}_\text{lpips}(\hat{A},A)\\
& + \mathcal{L}_\text{lpips}(\hat{A_{r}},A_{r})
+ \lambda_g ||\hat{T_{g}}-T_{g}||_{1},%
\end{aligned}%
\end{equation}%
where $\mathcal{L}_\text{lpips}$ denotes a perceptual loss~\cite{zhang2018perceptual}, $\hat{A_{r}}$, $\hat{A}$, and $\hat{T_{g}}$ are the {rendered} albedo images in the raw and super-resolution domains, and the predicted {albedo} tri-plane, respectively. $A$, $A_{r}$, and $T_{g}$ are the corresponding ground truth. The parameter $\lambda_g$ decreases from $1$ to $0.01$ after the initial 8 million iterations.
\paragraph{Shading Loss.} This loss measures the disparity between the predicted and ground-truth shading features. It is defined as
\begin{equation}
\label{equation:shading loss}
\mathcal{L}_\text{shading} = ||\hat{S}-S||_{1}
+ \lambda_s ||\hat{T_{S}}-T_{S}||_{1}  ,
\end{equation}%
where $\hat{S}$ and $\hat{T_{S}}$ are the predicted shading maps and the shading tri-plane, respectively, and $S$ and $T_{S}$ are the corresponding ground truth. The parameter $\lambda_s$ decreases from $1$ to $0.01$ after the initial 8 million iterations.
\paragraph{RGB Loss.} This loss assesses the dissimilarity between the predicted and ground-truth composed images in the raw, super-resolution, and feature domains. In addition to a perceptual loss~\cite{zhang2018perceptual}, an identity loss~\cite{ArcFace} is employed to retain the appearance and identity of facial images. The RGB loss is defined as
\begin{equation}
\label{equation:rgb loss}
\begin{aligned}
\mathcal{L}_\text{rgb} & =  ||\hat{I}-I||_{1}
+ ||\hat{I_{r}}-I_{r}||_{1}
+ \mathcal{L}_\text{lpips}(\hat{I},I) \\
&+ \mathcal{L}_\text{lpips}(\hat{I_{r}},I_{r})
+ \lambda_f ||\hat{I_{f}}-I_{f}||_{1}
+ \mathcal{L}_{id}(\hat{I},I) ,
\end{aligned}%
\end{equation}%
where $\hat{I}$, $\hat{I_{r}}$, and $\hat{I_{f}}$ are the predicted RGB images in the raw, super-resolution, and feature domains, respectively, and $I$, $I_{r}$, and $I_{f}$ are the corresponding ground truth. The parameter $\lambda_f$ decreases from $1$ to $0$ after the initial 8 million iterations.
\paragraph{Adversarial Loss.} This loss enforces the indistinguishability of the predicted RGB images from the source RGB images in both the raw and super-resolution domains. A dual discriminator $D$ from \cite{nerffacelighting} is utilized to discriminate between the predicted and real images. The adversarial loss is defined as
\begin{equation}
\label{equation:adv loss}
\begin{aligned}
\mathcal{L}_\text{adv} &= -(\mathbb E[\log D(I)] + 
\mathbb E[\log D(I_r)] \\
&+
\mathbb E[\log (1-D(\hat I))] +
\mathbb E[\log (1-D(\hat I_r))]).
\end{aligned}
\end{equation}

Our final loss function for training the dual-encoders is the weighted sum of the above losses:
\begin{equation}
\label{equation:total loss}
\begin{aligned}
    \mathcal{L} &= \lambda_\text{albedo} \mathcal{L}_\text{albedo} + \lambda_\text{shading} \mathcal{L}_\text{shading} \\
    &+\lambda_\text{rgb} \mathcal{L}_\text{rgb}
    + \lambda_\text{adv} \mathcal{L}_\text{adv} ,
\end{aligned}
\end{equation}
where $\lambda_\text{albedo}$, $\lambda_\text{shading}$, $\lambda_\text{rgb}$ and $\lambda_\text{adv}$ are the weights for each loss term. Initially, we set $\lambda_\text{albedo}=\lambda_\text{shading}=\lambda_\text{rgb}=1$ and $\lambda_\text{adv}=0$. After the first 16M iterations, we activate the adversarial loss by setting $\lambda_\text{adv}=1$ and keep the other weights unchanged. 

For training our temporal consistency network, {besides a reconstruction loss, }we use %
an additional temporal loss similar to \cite{chandran2022temporally,Lai-ECCV-2018} {to ensure }%
consistency in both short-term and long-term contexts. {Specifically, this loss is defined as follows:} 
\paragraph{Temporal Consistency Loss.}
{Without loss of generality, we assume current frame index is $i$ for discussion. }
The short-term temporal loss is computed by calculating the optical flow \( f_s \) between consecutive input frames \( F_i \) and \( F_{i-1} \). %
Subsequently, the previous outputs are warped to align with the current frame. Formally, the short-term temporal loss is defined as:
\begin{equation}
\label{equation:short-term temporal loss}
\mathcal{L}_\text{short} = M_{s}^{i}  \sum_{\omega \in \{\hat{I}, \hat{I}_r, \hat{A}, \hat{A}_r, \hat{S}\} } \mathcal{L}_\text{lpips}(\omega^i - \widetilde{\omega}^{i-1}) ,
\end{equation}%
where \( \hat{I}^i, \hat{I}_r^i, \hat{A}^i, \hat{A}_r^i, \) and \( \hat{S}^i \) represent the currently predicted RGB image, raw RGB image, albedo image, raw albedo image, and shading image, {based on the summation of original tri-planes and predicted residual tri-planes, }respectively. Similarly, \( \widetilde{I}^{i-1}, \widetilde{I}_r^{i-1}, \widetilde{A}^{i-1}, \widetilde{A}_r^{i-1}, \) and \( \widetilde{S}^{i-1} \) are the corresponding frames warped using \( f_s \) from the previous time step. The mask \( M_s^i \) is defined as \( M_s^i = \exp(||I^i - \widetilde{I}^{i-1}||_{1}) \), which mitigates errors introduced during the warping process. %

For the long-term temporal loss, the same procedure is applied, but with the temporal index \( i-1 \) replaced by 1. In other words, this process ensures temporal consistency between the first frame and the current frame. Similarly, the long-term temporal loss is defined as
\begin{equation}
\label{equation:long-term temporal loss}
\mathcal{L}_\text{long} = M_{l}^{i}  \sum_{\omega \in \{\hat{I}, \hat{I}_r, \hat{A}, \hat{A}_r, \hat{S}\} } \mathcal{L}_\text{lpips}(\omega^i - \widetilde{\omega}^{1}),
\end{equation}%
where \( \widetilde{I}^{i-1}, \widetilde{I}_r^{i-1}, \widetilde{A}^{i-1}, \widetilde{A}_r^{i-1}, \) and \( \widetilde{S}^{i-1} \) are the corresponding frames warped using \( f_l \) from the first time step. The mask \( M_l^i \) is defined as \( M_l^i = \exp(||I^i - \widetilde{I}^{1}||_{1}) \). 

Our final loss function for training the temporal consistency network is
\begin{equation}
\label{equation:temporal loss}
\mathcal{L}_\text{temporal} = \lambda_\text{short}\mathcal{L}_\text{short} + \lambda_\text{long}\mathcal{L}_\text{long} + \lambda_\text{lpips}\mathcal{L}_{\text{lpips}}(\hat{I}^i,I^i).
\end{equation}
{where $I^i$ denotes the ground-truth image,  $\mathcal{L}_{\text{lpips}}$ promotes the reconstruction and $\lambda_\text{short}=1,\lambda_\text{long}=1,\lambda_\text{lpips}=1$}.

\section{Experiments}
\label{sec:Experiments}
\newcommand{\pasteSH}[1]{\llap{\includegraphics[width=0.03\textwidth]{#1}}}

\begin{figure*}[t] %
\def\lw{0.132\linewidth}
\def\hlw{0.066\lw}
\def\ftsz{\normalsize}
\renewcommand\tabcolsep{0.0pt}
\renewcommand{\arraystretch}{0}
\centering \small
\begin{tabular}{ccccccc}
\includegraphics[width=\lw]{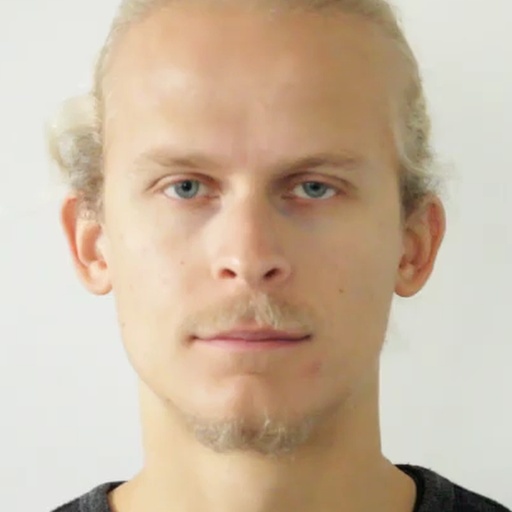} &
\includegraphics[width=\lw]{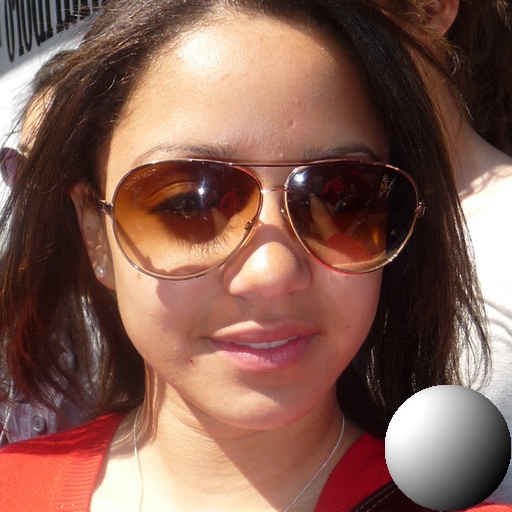}&
\includegraphics[width=\lw]{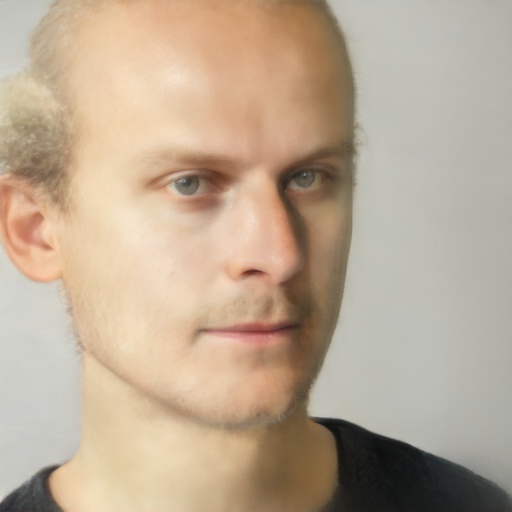} &
\includegraphics[width=\lw]{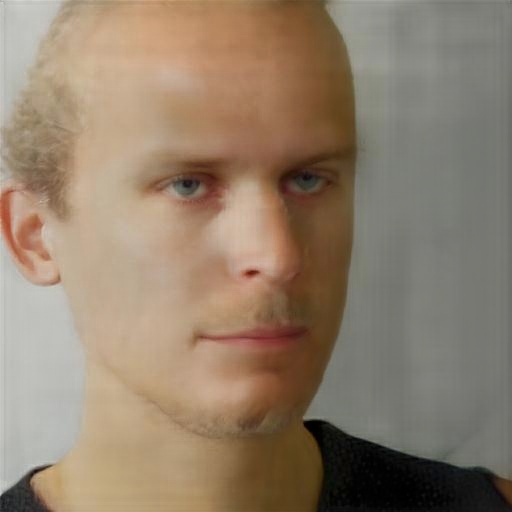} &
\includegraphics[width=\lw]{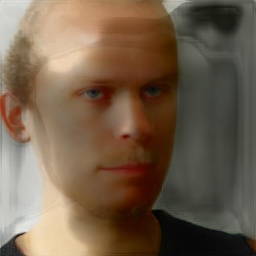} &
\includegraphics[width=\lw]{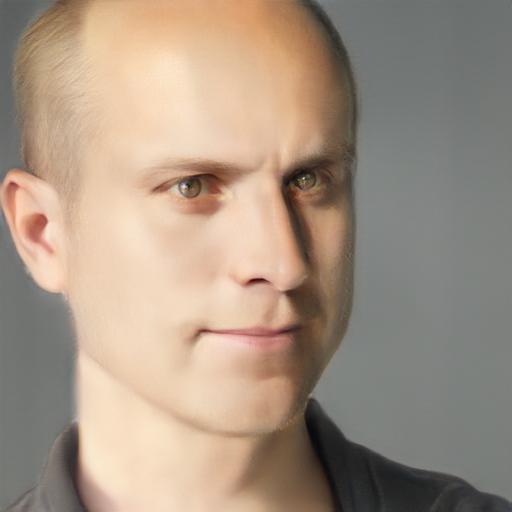} &
\includegraphics[width=\lw]{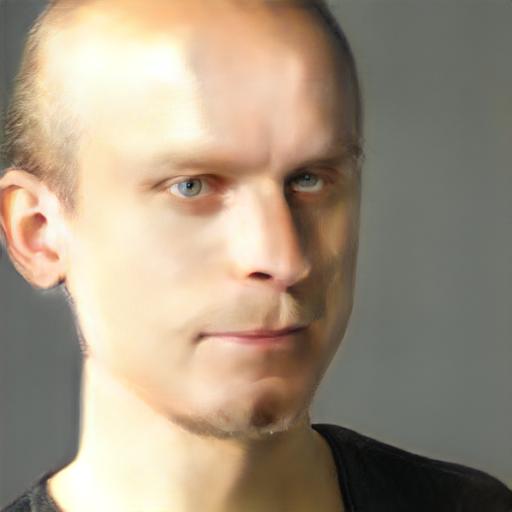} \\
\includegraphics[width=\lw]{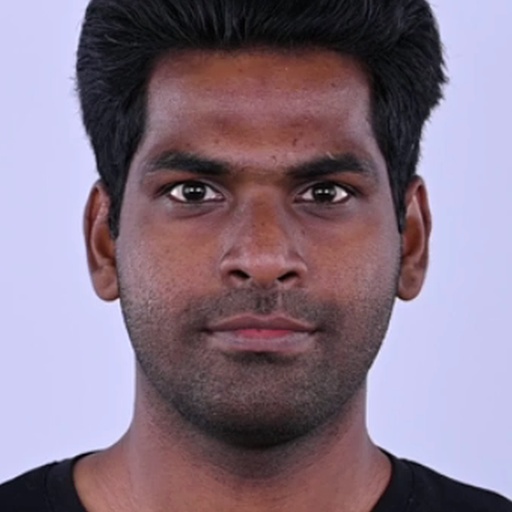} &
\includegraphics[width=\lw]{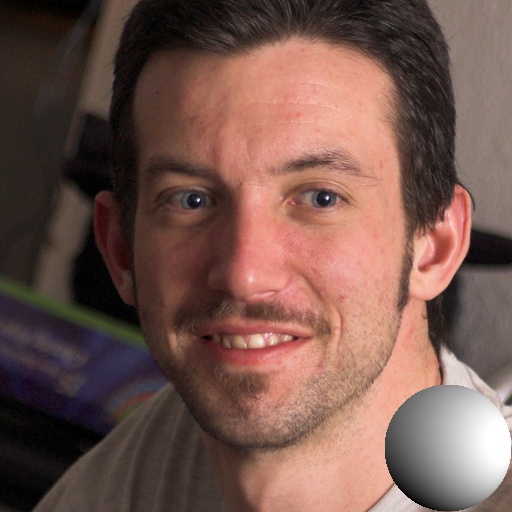} &
\includegraphics[width=\lw]{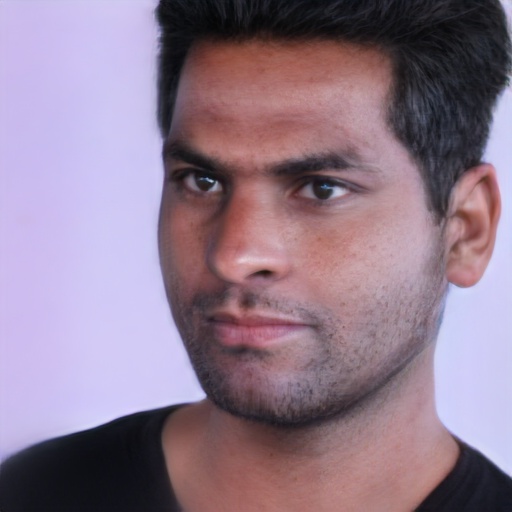} &
\includegraphics[width=\lw]{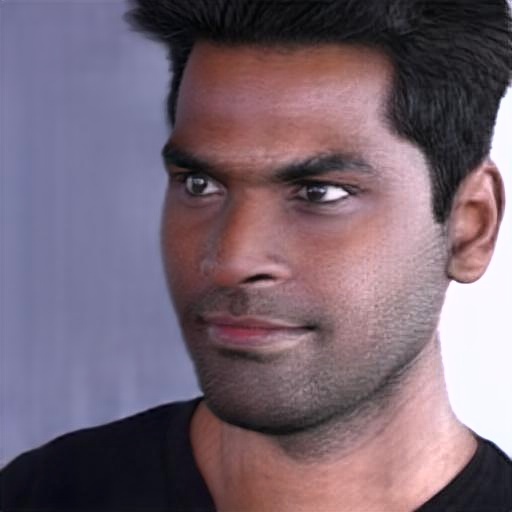} &
\includegraphics[width=\lw]{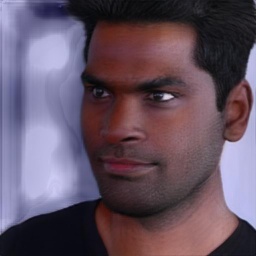} &
\includegraphics[width=\lw]{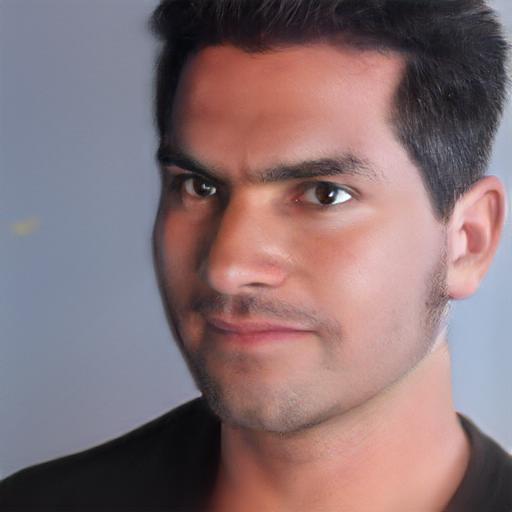} &
\includegraphics[width=\lw]{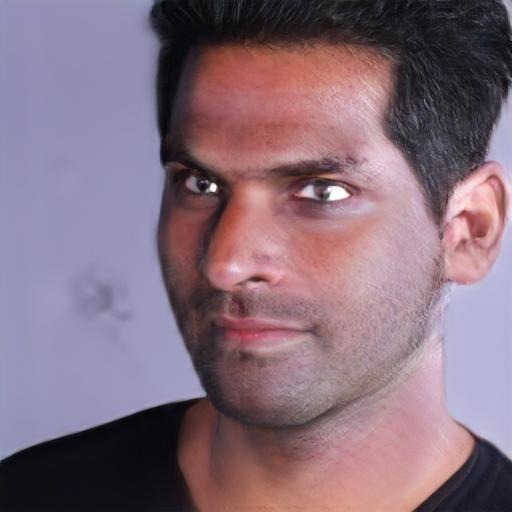} \\
\vspace{0.3mm}
\\
Input & Light Reference & Ours & B-DPR & B-SMFR & B-E4E & B-PTI 

\end{tabular}
\caption{Comparison of video relighting quality on novel views. Our method produces more realistic and consistent results than the baseline methods introduced in \sref{subsec:Quantitative Evaluation}.}
\label{fig:novel view relighting} 
\end{figure*}

\begin{figure}[htbp] 
\def\lw{0.22\linewidth}
\def\hlw{0.11\lw} 
\renewcommand\tabcolsep{0.0pt}
\renewcommand{\arraystretch}{0}
\centering \small
\begin{tabular}{ccccc} 
 \rotatebox{90}{\hspace{5.5mm}Input} &
\includegraphics[width=\lw]{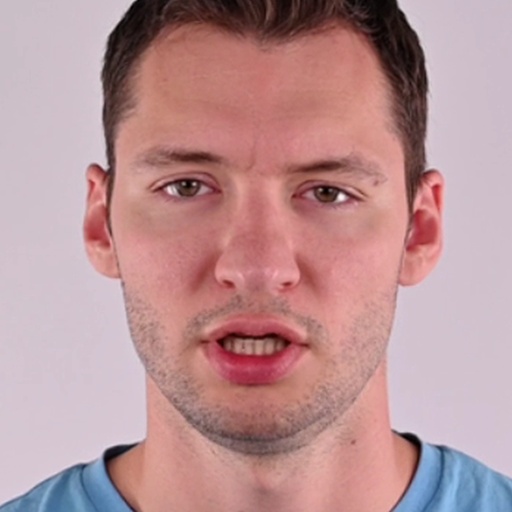} & 
\includegraphics[width=\lw]{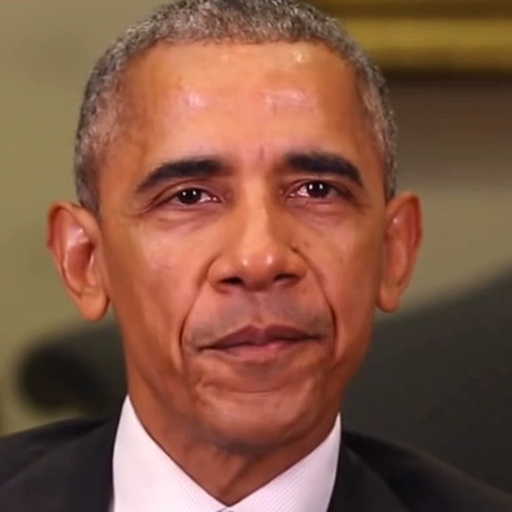} & 
\includegraphics[width=\lw]{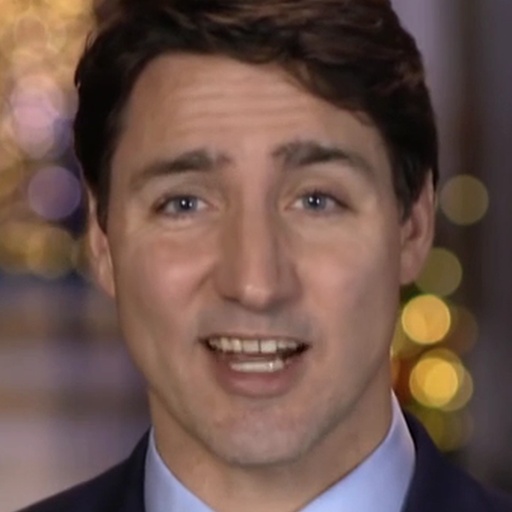} & 
\includegraphics[width=\lw]{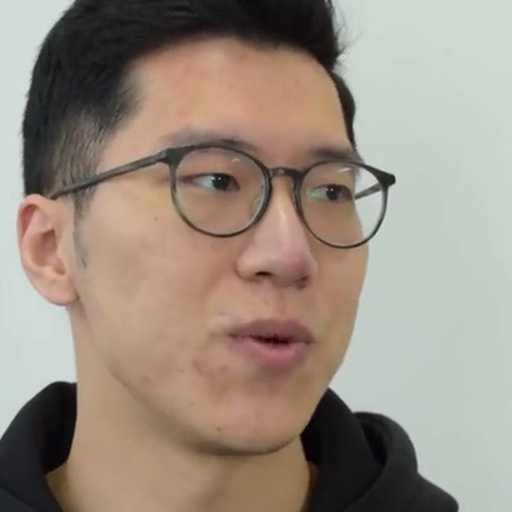} 
 \\
 \rotatebox{90}{\hspace{3.5mm}Reference}  &
 \includegraphics[width=\lw]{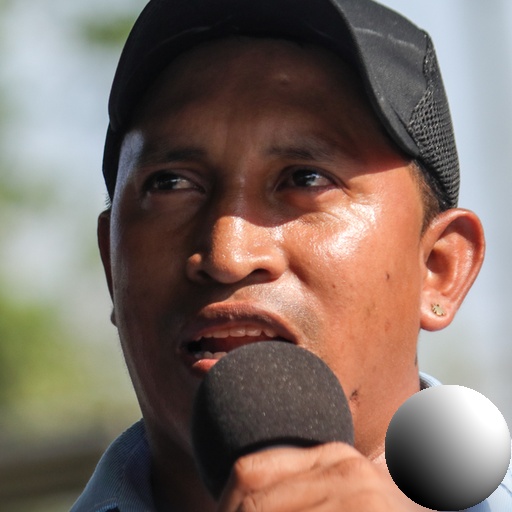} & 
 \includegraphics[width=\lw]{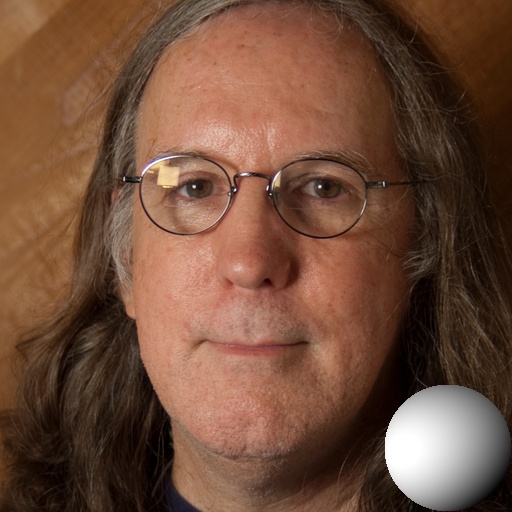} & 
\includegraphics[width=\lw]{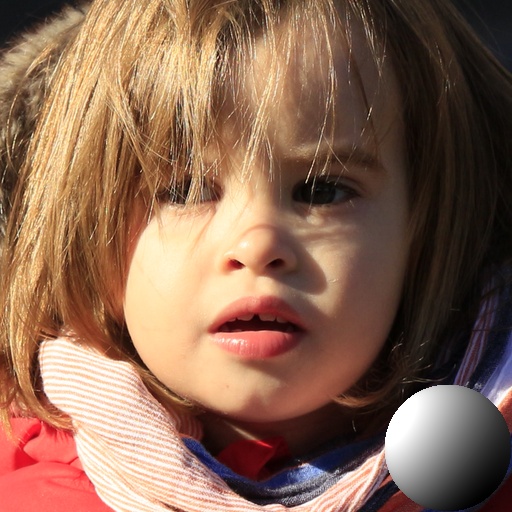} & 
\includegraphics[width=\lw]{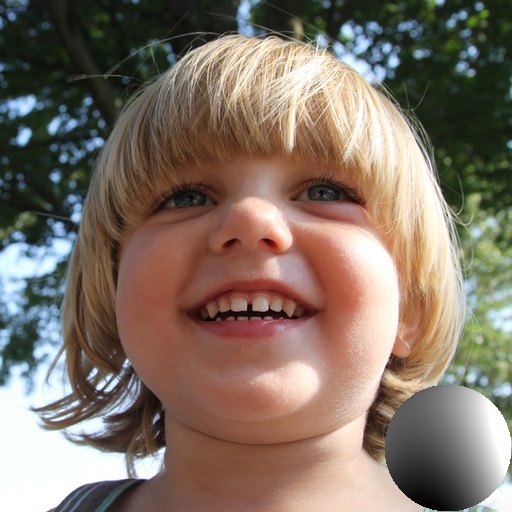}
 \\
 \rotatebox{90}{\hspace{3mm}SMFR~\cite{SMFR}}  &
  \includegraphics[width=\lw]{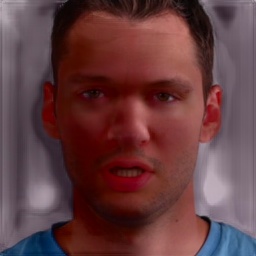}  & 
 \includegraphics[width=\lw]{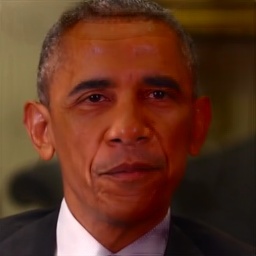} & 
\includegraphics[width=\lw]{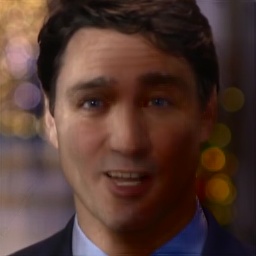} & 
\includegraphics[width=\lw]{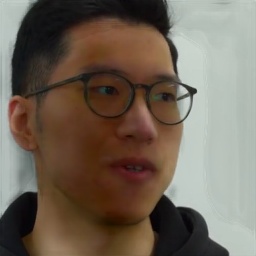} 
 \\
  \rotatebox{90}{\hspace{4mm}DPR~\cite{DPR}} &
  \includegraphics[width=\lw]{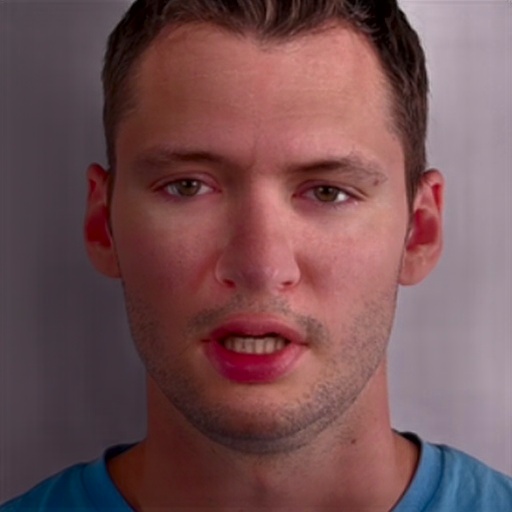} & 
 \includegraphics[width=\lw]{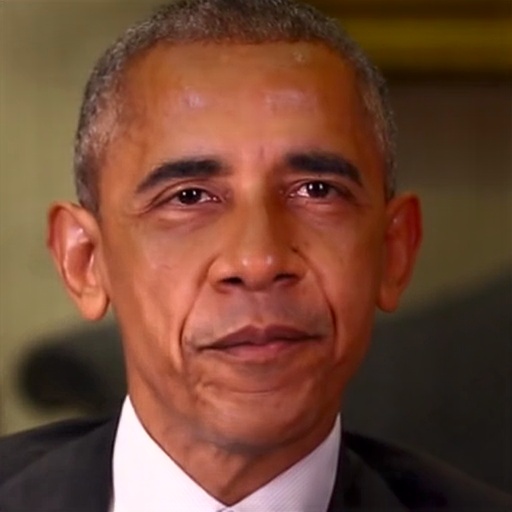} & 
\includegraphics[width=\lw]{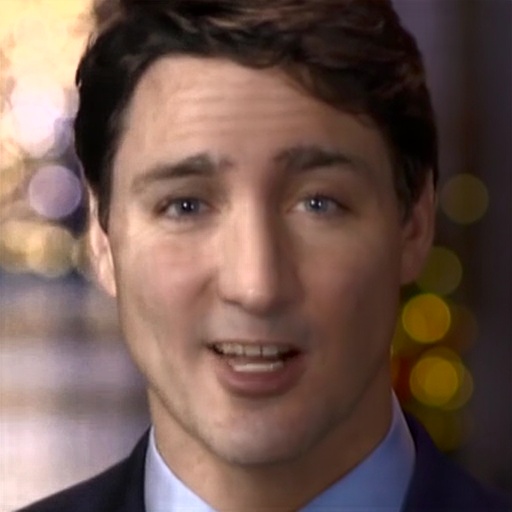} & 
\includegraphics[width=\lw]{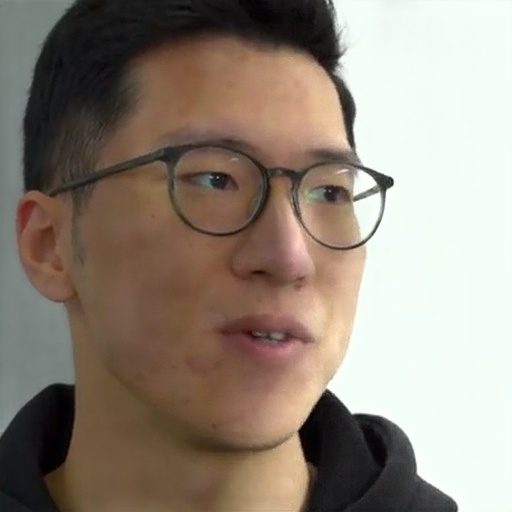}
 \\
\rotatebox{90}{\hspace{1mm}ReliTalk~\cite{qiu2023relitalk}} &
\includegraphics[width=\lw]{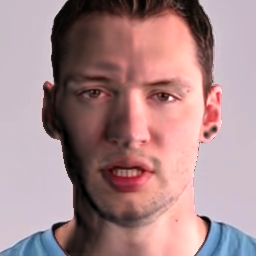} & 
\includegraphics[width=\lw]{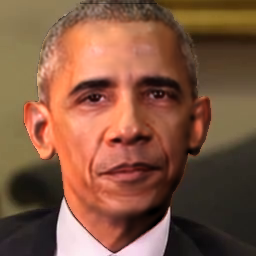} & 
\includegraphics[width=\lw]{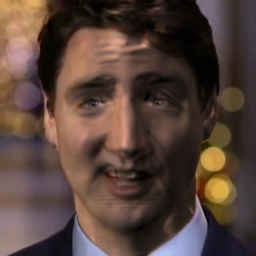} & 
\includegraphics[width=\lw]{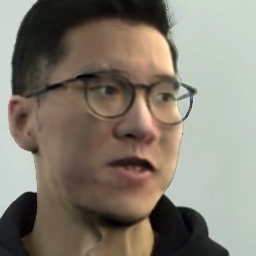} 
 \\
\rotatebox{90}{\hspace{7mm}Ours} &
\includegraphics[width=\lw]{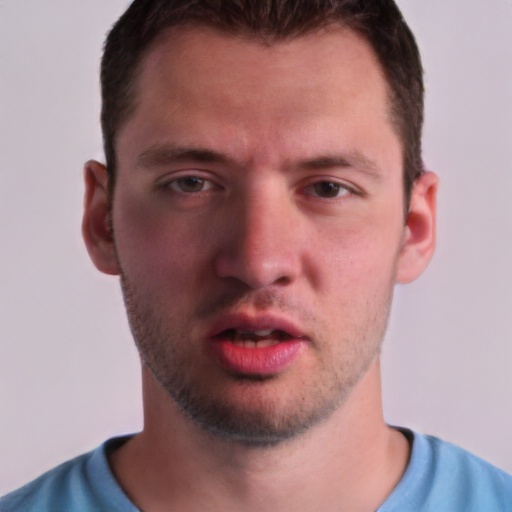} & 
\includegraphics[width=\lw]{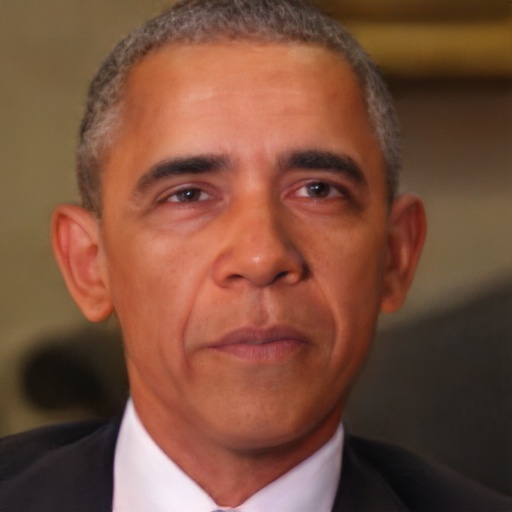} &
\includegraphics[width=\lw]{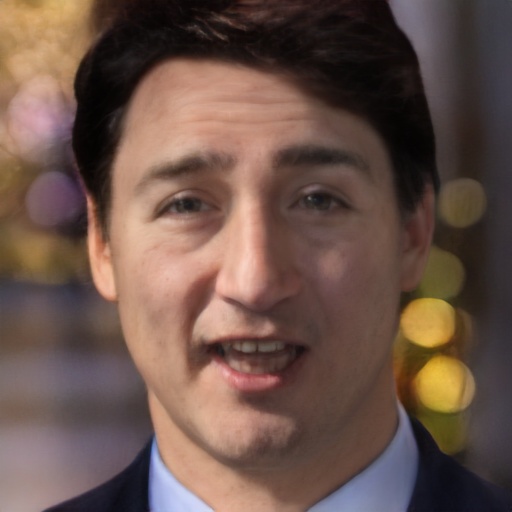} & 
\includegraphics[width=\lw]{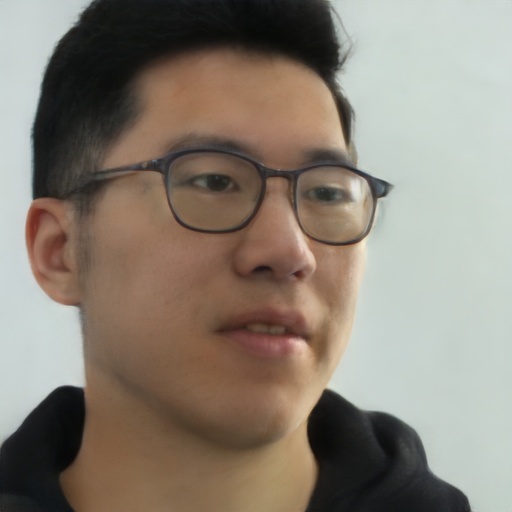} 
\end{tabular} 
\caption{Comparison of video relighting quality in the input view. We compare our method with three methods: SMFR~\cite{SMFR}, DPR~\cite{DPR}, and ReliTalk~\cite{qiu2023relitalk}. We show the input video frames in the first row and the relighted results under different lighting conditions in the remaining rows. Our method produces more realistic and consistent results than other methods, especially under challenging conditions like the side lighting.} \label{fig:input view relighting} 
\vspace{-1mm}
\end{figure}

\begin{figure*}[t] 
\def\lw{0.111\linewidth}
\def\hlw{0.0555\linewidth}
\def\ftsz{\normalsize}
\renewcommand\tabcolsep{0.0pt}
\renewcommand{\arraystretch}{0}
\centering \small
\begin{tabular}{ccccccccc}
\includegraphics[width=\lw]{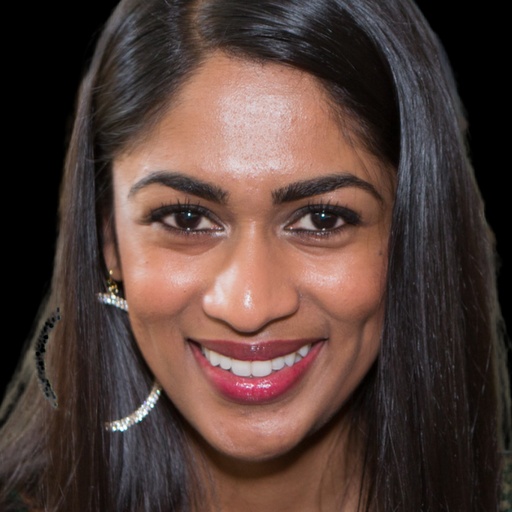} 
& \includegraphics[width=\hlw]{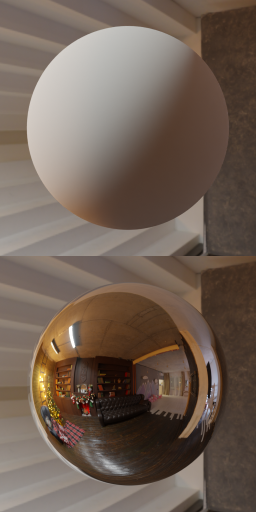}  
& \includegraphics[width=\lw]{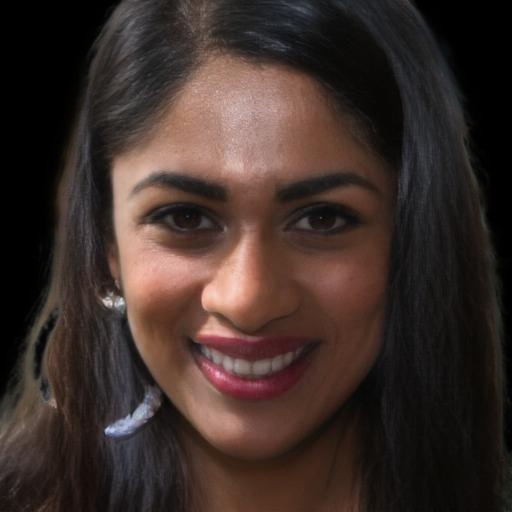}  
& \includegraphics[width=\lw]{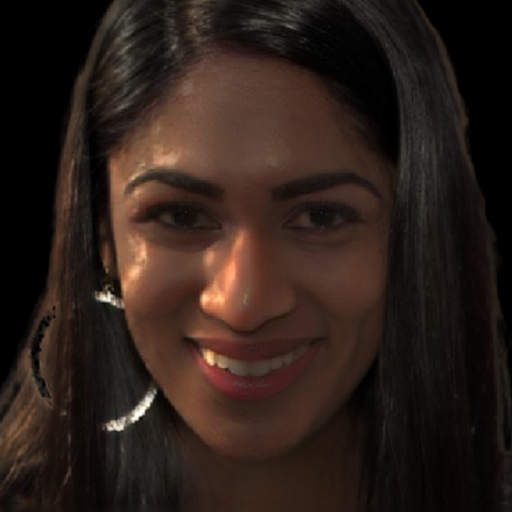}  
& \includegraphics[width=\lw]{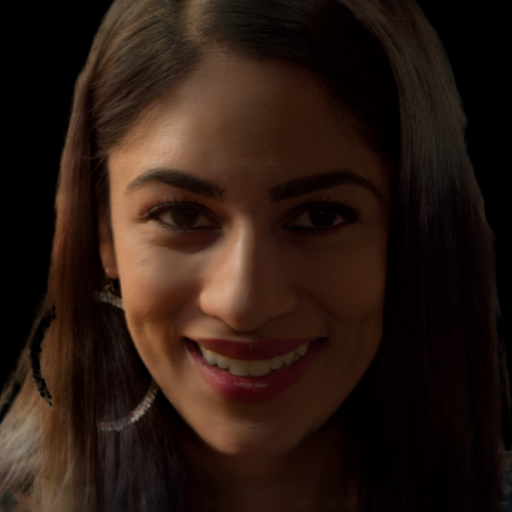}  
& \includegraphics[width=\lw]{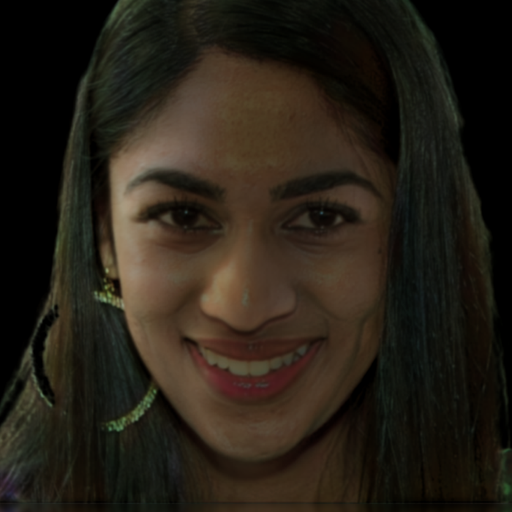}  
& \includegraphics[width=\lw]{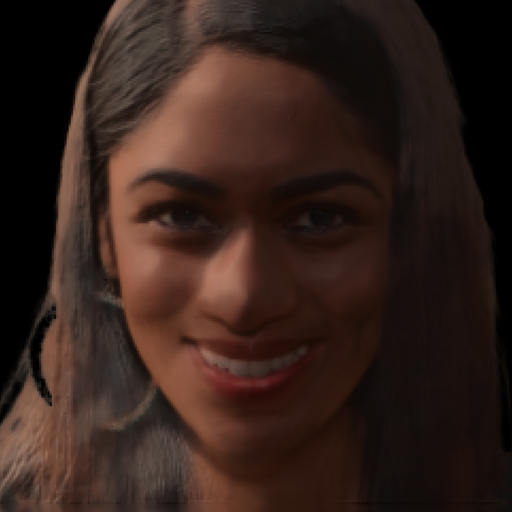} 
& \includegraphics[width=\lw]{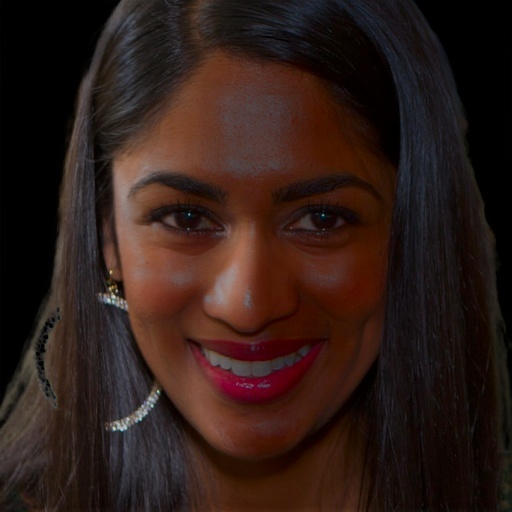}  
& \includegraphics[width=\lw]{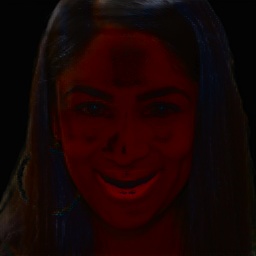} \\
\includegraphics[width=\lw]{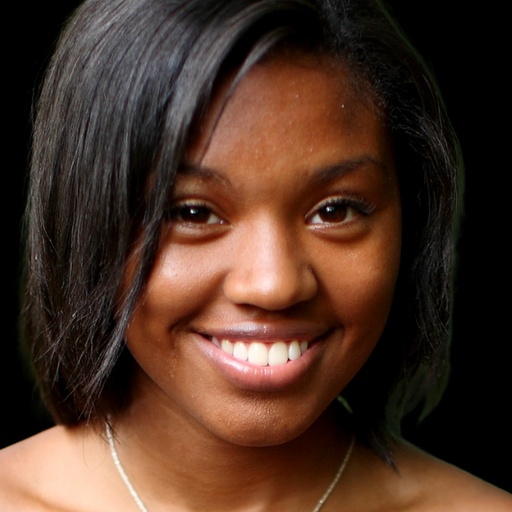} 
& \includegraphics[width=\hlw]{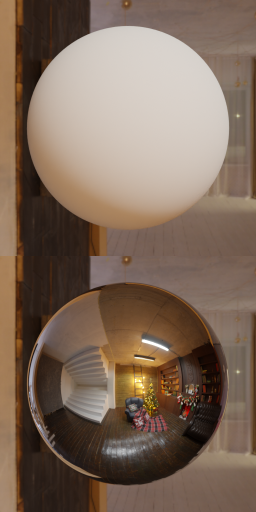}  
& \includegraphics[width=\lw]{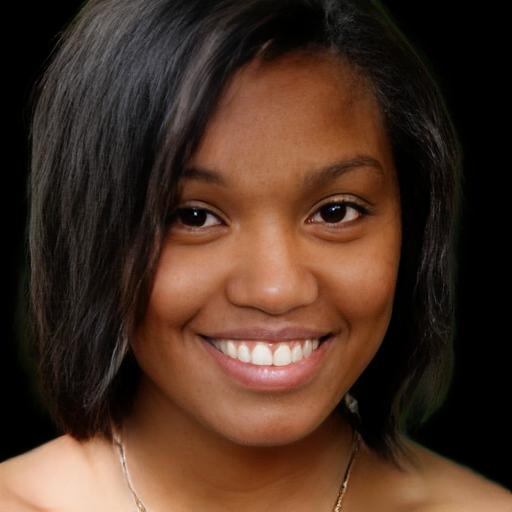}  
& \includegraphics[width=\lw]{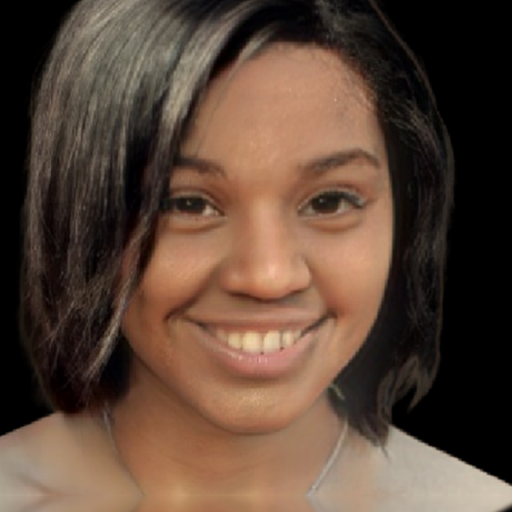}  
& \includegraphics[width=\lw]{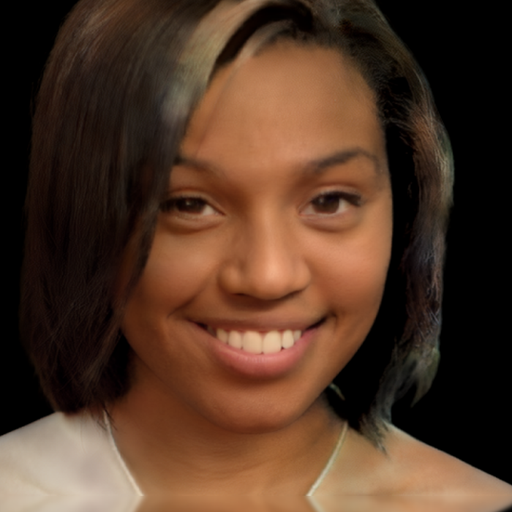}  
& \includegraphics[width=\lw]{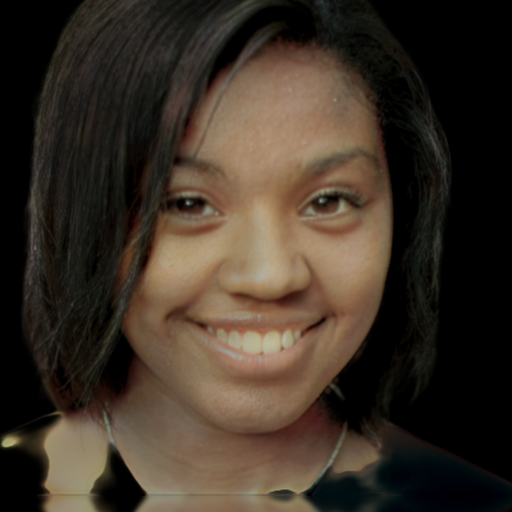}  
& \includegraphics[width=\lw]{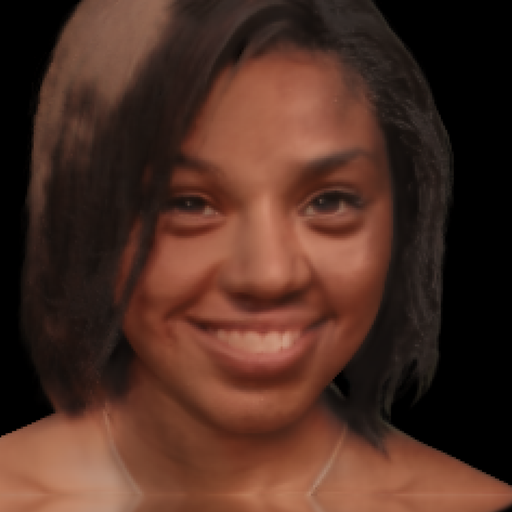}
& \includegraphics[width=\lw]{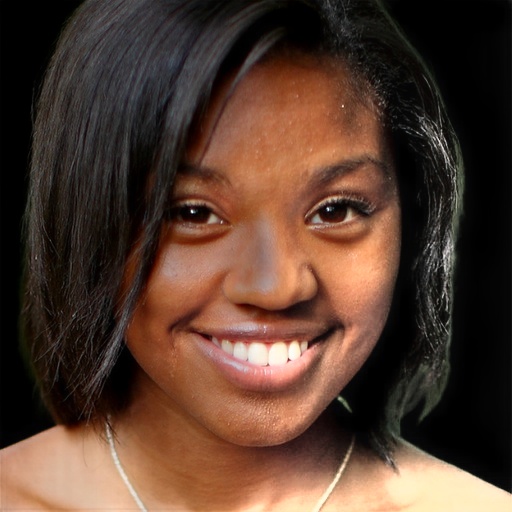}  
& \includegraphics[width=\lw]{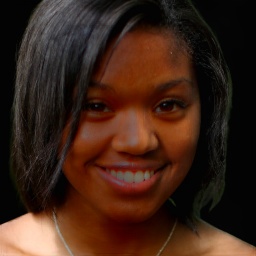}\\
\includegraphics[width=\lw]{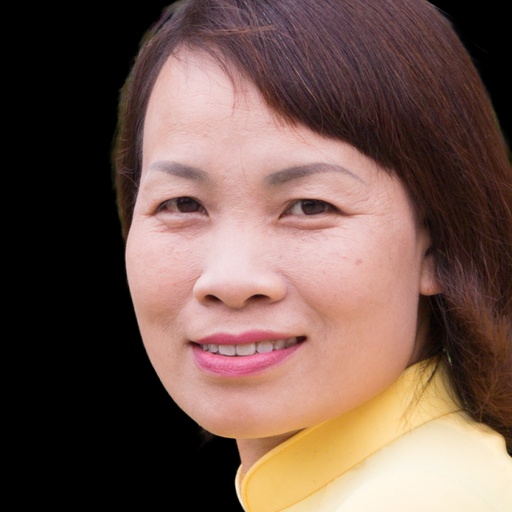} 
& \includegraphics[width=\hlw]{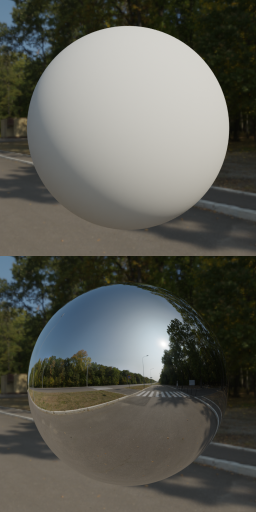} 
& \includegraphics[width=\lw]{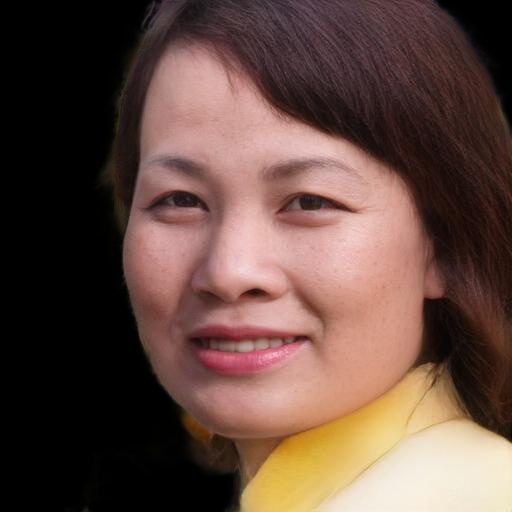}  
& \includegraphics[width=\lw]{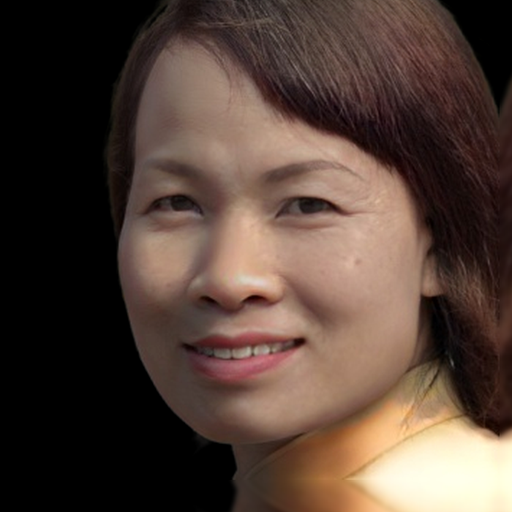}  
& \includegraphics[width=\lw]{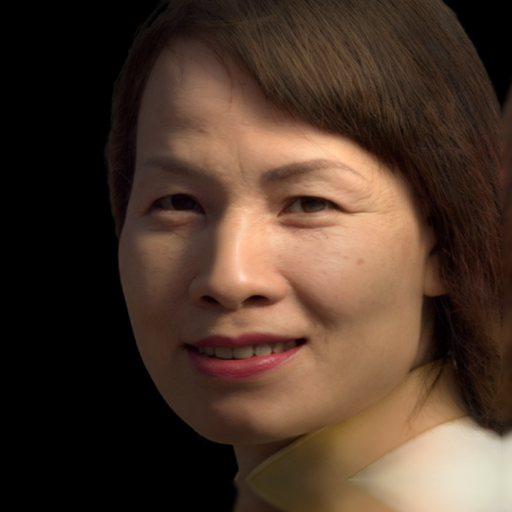}  
& \includegraphics[width=\lw]{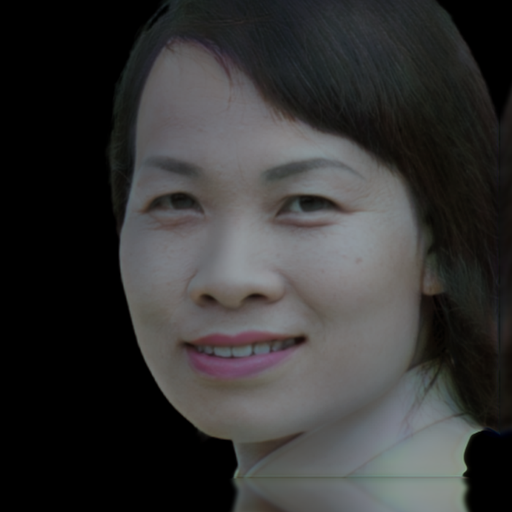}  
& \includegraphics[width=\lw]{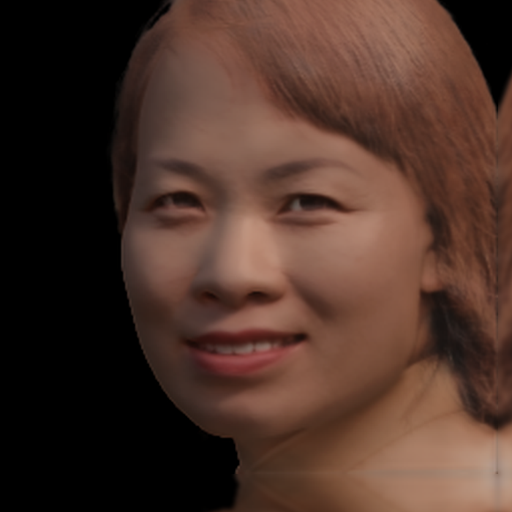}
& \includegraphics[width=\lw]{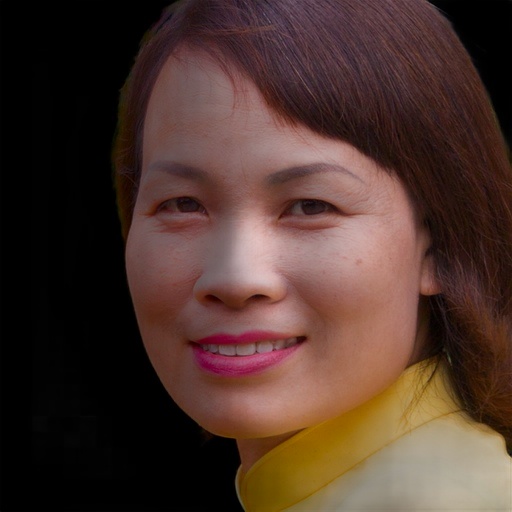}  
& \includegraphics[width=\lw]{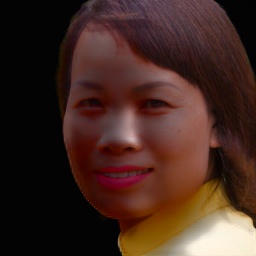}\\
\vspace{0.5mm}
\\
\ftsz{Input} & \ftsz{Light} & \ftsz{Ours} & \ftsz{Lumos~\cite{yeh2022learning}} & \ftsz{TR~\cite{pandey2021total}} & \ftsz{NVPR~\cite{Zhang_2021_ICCV}} & \ftsz{SIPR-W~\cite{wang2020single}} & \ftsz{DPR~\cite{DPR}} & \ftsz{SMFR~\cite{SMFR}} 
\end{tabular}

\caption{
Comparison of relighting quality on the input view. We compare our method with six methods: Lumos~\cite{yeh2022learning}, TR~\cite{pandey2021total}, NVPR~\cite{Zhang_2021_ICCV}, SIPR-W~\cite{wang2020single}, DPR~\cite{DPR} and SMFR~\cite{SMFR}. We show the input image in the first column, the sphere renderings from the environment map in the second column, and the relighted results in the remaining columns. Our method produces more realistic and consistent results than the other methods.} 
\label{fig:lumos relighting} 
\end{figure*}

In this section, we show our experimental setup and discuss the results of our experiments. 
The comparisons with alternative methods, and ablation study 
show the effectiveness of our method and its superiority to the alternative approaches. 

\subsection{Implementation Details}
\label{subsec:Implementation Details}

\paragraph{Datasets.}
\label{subsubsec:Datasets}
We evaluate our method on the portrait videos from INSTA~\cite{INSTA:CVPR2023}, which consist of 31,079 frames in total. Following \cite{pv3d}, we crop the images and videos to focus on the faces. We estimate the camera pose for each frame using the technique from \cite{chan2022efficient}. We also extract the lighting conditions with DPR~\cite{DPR}.

\paragraph{Training Details.}
\label{subsubsec:Training Details}
{As to the tri-plane dual-encoders, }
we first freeze the generator and train only our encoder. After the first 16M iterations, we unfreeze the albedo decoder, shading decoder, and super-resolution module and train them jointly with the dual-encoders. 
{As to the temporal consistency network, }
we sample camera poses %
from %
normal and uniform distributions for each person. We use two views for each person. For the first view, we sample the focal length, camera radius, principal point, camera pitch, camera yaw, and camera roll from $\mathrm{N}(18.837, 1)$, $\mathrm{N}(2.7, 0.1)$, $\mathrm{N}(256, 14)$, $\mathrm{U}(-26^\circ,26^\circ)$, $\mathrm{U}(-49^\circ,49^\circ)$, and $\mathrm{N}(0, 2^\circ)$, respectively. For the second view, we sample the camera pitch and camera yaw from $\mathrm{U}(-26^\circ,26^\circ)$ and $\mathrm{U}(-36^\circ,36^\circ)$, respectively and fix the other parameters to 18.837 (focal length), 2.7 (camera radius), 256 (principal point), and 0 (camera roll). 

We train our network using the Adam~\cite{kingma2014adam} optimizer with a learning rate of 0.0001, except for the Transformer parameters, which have a learning rate of 0.00005. It takes about 30 days to train our network on 8 NVIDIA Tesla V100 GPUs with batch size 32. More details can be found in the supplementary material.

{\paragraph{Inference Speed.} We employ a single RTX 4090 GPU during inference. The average inference time for each frame is 30.32 milliseconds, resulting in an average of 32.98 frames per second (fps), excluding secondary tasks such as image I/O and data transfer between the CPU and GPU.}
\subsection{Quantitative Evaluation}
\label{subsec:Quantitative Evaluation}

To evaluate the performance of our method, we compare it with other methods capable of 3D-aware portrait relighting. However, none of existing techniques can achieve this goal in a single step, so we have to combine different methods to construct the baselines. Specifically, we use the following methods. 
\textbf{{B-DPR}} uses PTI~\cite{roich2022pivotal} to invert each frame of an input video as a latent code of EG3D~\cite{chan2022efficient}, allowing for rendering novel views and relighting using DPR~\cite{DPR}.
\textbf{{B-SMFR}} uses the same inversion and rendering method as {B-DPR}, but uses SMFR~\cite{SMFR} to relight the rendered frames from novel views.
\textbf{{B-E4E}} uses an off-the-shelf encoder from a state-of-the-art NeRF-based face image relighting method~\cite{nerffacelighting} to invert each frame of the input video and relight it from novel views, which achieves real-time performance at the cost of quality.
\textbf{{B-PTI}} uses the same encoder as {B-E4E}, but we apply the PTI~\cite{roich2022pivotal} to fine-tune a single generator for each input video. This improves the reconstruction quality but takes more {training} time than {B-E4E}. 
We evaluate the performance of different methods regarding reconstruction quality, novel view relighting quality, identity perseverance, and time cost.

\paragraph{Novel View Relighting Quality.}
\label{subsubsec:Novel View Relighting Quality}

To evaluate the relighting quality under novel views, we 
{relight first} 500 frames from each video from \cite{INSTA:CVPR2023}. We render each video from three novel views and pair them with five distinct lighting conditions, resulting in a total of 75,000 frames for a comprehensive comparison. Following \cite{nerffacelighting}, we adopt an off-the-shelf estimator \cite{DECA:Siggraph2021} to calculate the lighting accuracy and {instability}. We use MagFace~\cite{meng2021magface}, different from the one we use in training, to measure identity preservation between different views. {To assess temporal consistency, we use an optical flow estimator~\cite{teed2020raft} to calculate warping error (WE). This involves warping the preceding frame to align with the current frame and measuring MSE loss. We also compute the LPIPS between adjacent frames for an additional evaluation of temporal consistency.} We list the time each method takes to relight a face.
\Tref{tab:Novel View Relighting Quality} summarizes the quantitative evaluation results using the lighting error, lighting {instability} {calculated based on the lighting transfer task introduced in \cite{nerffacelighting}}%
, identity preservation (ID), and processing time (Time) on the INSTA~\cite{INSTA:CVPR2023} video dataset. Our method outperforms the baselines, demonstrating the second lowest lighting error and {instability}, the highest identity preservation, the lowest warping error and LPIPS, and the lowest time cost. 

\begin{table}[t]
\vspace{1mm}
\centering
\caption{Quantitative evaluation using lighting error (LE), lighting instability (LI), Identity Perservance (ID), Warping Error (WE)%
, LPIPS between consecutive frames and avarage time cost (Time)
on the INSTA~\cite{INSTA:CVPR2023} video dataset. {We highlight the best score in boldface and underline the second best.}} 
\label{tab:Novel View Relighting Quality}
\vspace{-1mm}
\fontsize{7pt}{8pt}\selectfont
\begin{tabular}{@{}lcccccc@{}}
\toprule
&LE$\downarrow$&LI$\downarrow$& ID$\uparrow$ & WE$\downarrow$& LPIPS$\downarrow$&Time (s) $\downarrow$ \\ \midrule

B-DPR      &  0.9093 &  0.3041&  \underline{0.5222}& 0.0029&0.1015&200\\
B-SMFR    &     1.0929&  0.3352&  0.4479& 0.0022&0.0626&200\\
B-E4E    &  \textbf{0.6384}   & \textbf{0.1963} &  0.2892& \underline{0.0007}&\underline{0.0306}&\underline{0.2}\\
B-PTI    & 0.8220    & 0.2630 & 0.4728 & 0.0049& 0.1080&30\\
Ours    &  \underline{0.7710} & \underline{0.2533} &  \textbf{0.5396}& \textbf{0.0003}&\textbf{0.0159}&\textbf{0.03}\\ \bottomrule
\end{tabular}
\end{table}

\paragraph{Reconstruction Quality.}
\label{subsubsec:Reconstruction Quality}

To assess the quality of reconstruction, we use four quantitative metrics: LPIPS~\cite{zhang2018perceptual}, DISTS~\cite{ding2020dists}, Pose Error (Pose), and Identity Preservation (ID). We obtained {and used} the same test data as LP3D~\cite{LP3D}.%

\begin{table}[t]
\centering
\caption{Quantitative evaluation using LPIPS, DISTS, Pose Accuracy (Pose), and Identity Consistency (ID) on 500 FFHQ images. $^\dagger$Evaluated only using the face region. $^\ddagger$Evaluated only using the foreground on $256^2$ images. We highlight the best score in boldface and underline the second best. 
}
\label{tab:reconstruction}
\vspace{-1mm}
\fontsize{7pt}{8pt}\selectfont
\begin{tabular}{@{}lcccc@{}}
\toprule
        &  LPIPS$\downarrow$  & DISTS$\downarrow$   & Pose$\downarrow$  & ID$\uparrow$    \\ \midrule
HeadNeRF$^\dagger$  & .2502 & .2427  & .0644 & .2031 \\
LP3D$^\dagger$     & \textbf{.1240} & \textbf{.0770}  & \underline{.0490} & .5481 \\
Ours$^\dagger$     &  \underline{.1746}    &  \underline{.1134}      &  \textbf{.0323 }    &  .\textbf{7109 }    \\ \midrule
ROME$^\ddagger$     & .1158 & .1058  & .0637 & .3231 \\
LP3D$^\ddagger$     & .\textbf{0468} & \textbf{.0407}  & \underline{.0486} & \underline{.5410} \\
Ours$^\ddagger$     &  \underline{.1053}     &  \underline{.0835}      & \textbf{.0327 }     &   \textbf{.7201}    \\ \midrule
EG3D-PTI & .3236 & .1277  & .0575 & .4650 \\
LP3D     & .2692 & .\textbf{0904}  & .0485 & .5426 \\
LP3D(LT) & .2750 & \underline{.1021}  & .0448 & .5404 \\
NFL-PTI & \textbf{.2332} & .1627  & \textbf{.0228 }& \underline{.6825} \\
Ours     & \underline{.2400} & .1282  &  \underline{.0365}     & \textbf{.7015 }\\ \bottomrule
\end{tabular}
\vspace{-1.5mm}
\end{table}

\paragraph{Input View Relighting Quality.}
\label{subsubsecd:Relighting Quality}

We compare our method with four state-of-the-art portrait relighting methods: SIPR-W~\cite{wang2020single}, TR~\cite{pandey2021total}, NVPR~\cite{Zhang_2021_ICCV}, and Lumos~\cite{yeh2022learning}. %
We follow the same protocol as Lumos to obtain the results for comparison. 
As shown in \Tref{tab:input view relighting}, our method achieves the lowest Fréchet Inception Distance, suggesting more realistic outcomes, and the highest Identity Preservation.
For a visual comparison, please refer to Figure \ref{fig:input view relighting}, where our approach yields the most realistic and natural results.

{For the video input,} we evaluate the relighting accuracy and {instability} while performing the video relighting on the input view. Following \cite{nerffacelighting}, we adopt an off-the-shelf estimator \cite{DECA:Siggraph2021}, which is different from the one~\cite{DPR} we use during the inference time, to calculate the lighting accuracy and the lighting {instability}. 
{As shown in \Tref{tab:insta video relighting}. Compared to DPR, SMFR and ReliTalk, our method achieves the lowest
lighting instability and the second lowest lighting error.}

\begin{table}[t]
\vspace{1mm}
\centering
\caption{Quantitative evaluation on the cropped test set of FFHQ~\cite{karras2019style}. We highlight the best score in boldface and underline the second best.} %
\fontsize{\tbfta}{\tbftb}\selectfont
\vspace{-1mm}
\begin{tabular}{@{}lcccccc@{}}
\toprule
& SIPR-W & NVPR & TR & Lumos & SMFR & Ours \\
\midrule
FID$\downarrow$ & 87.39 & 65.23 & 55.30 & 55.18 & \underline{51.16} & \textbf{45.08} \\
ID$\uparrow$ & 0.6442 & 0.7242 & 0.6193 & \underline{0.7374}  & 0.6285 &\textbf{0.7711} \\
\bottomrule
\end{tabular}
\label{tab:input view relighting}
\end{table}

\begin{table}[t]
\centering
\caption{Quantitative evaluation using the lighting error, lighting {instability}, {and average time cost (Time)} on the INSTA~\cite{INSTA:CVPR2023} video dataset. {We highlight the best score in boldface and underline the second best.} 
}%
\label{tab:insta video relighting}
\vspace{-1mm}
\fontsize{\tbfta}{\tbftb}\selectfont
\begin{tabular}{@{}lccc@{}}
\toprule

           &Lighting Error$\downarrow$&Lighting {Instability}$\downarrow$&Time~(s)$\downarrow$\\ \midrule

DPR~\cite{DPR}      & \textbf{0.7600}    & 0.2997 & \underline{0.04}\\
SMFR~\cite{SMFR}    & 1.1381    & \underline{0.2895} &  0.06\\
ReliTalk~\cite{qiu2023relitalk} & 1.2012     & 0.4060 & 0.20\\
Ours    & \underline{0.7816}  & \textbf{0.2841 }&  \textbf{0.03}\\ %

\bottomrule

\end{tabular}
\end{table}

\subsection{Qualitative Evaluation}
\label{subsec:Qualitative Evaluation}

We conduct a qualitative evaluation on portrait videos from \cite{Gafni_2021_CVPR_nerface} to demonstrate the effectiveness of our method.

\paragraph{Novel View Relighting Quality.} \Fref{fig:novel view relighting} shows our method's novel view synthesis capability under various viewpoints and lighting conditions. Among the five methods, our method preserves the lighting conditions of the reference images the most faithfully.

\paragraph{Input View Relighting Quality.} \Fref{fig:lumos relighting} presents the video relighting results in the input view by our method in comparison with three existing methods.
Our approach demonstrates superior accuracy in reproducing lighting effects, especially compared to existing non-3D-aware methods. This is particularly evident under challenging lighting conditions, such as side lighting, where our method outperforms others in maintaining image quality. 

\subsection{Ablation Study}
\label{subsec:Ablation Study}

We perform an ablation study to evaluate the necessity of each key component in our method.

\paragraph{Temporal Consistency Network.}
We remove the temporal consistency network and then compute lighting error and lighting instability based on the lighting transfer task introduced in \cite{nerffacelighting}. We also evaluate the temporal consistency based on the warp loss and LPIPS loss between consecutive frames, which serve as a reliable approximation of human perception regarding temporal consistency, capturing nuances like flickering effects. As shown in Table \ref{tab:ablation video relighting}, the absence of the temporal consistency network results in an increase in warping error and LPIPS, signaling a decline in temporal consistency.

\begin{table}[t]
\centering
\caption{Ablation study on temporal consistency network. We removed the temporal consistency network and calculate Lighting Error (LE), Lighting Instability (LI), Warping Error (WE) and LPIPS between consecutive frames. We highlight the best score in boldface.}
\label{tab:ablation video relighting}

\vspace{-1mm}
\fontsize{\tbfta}{\tbftb}\selectfont
\begin{tabular}{@{}lcccc@{}}
\toprule
 & LE$\downarrow$ &LI$\downarrow$&WE $\downarrow$ & LPIPS$\downarrow$\\ \midrule
 w/o TCN    & \textbf{0.7707}  & \textbf{0.2526} & 0.0006  & 0.0304 \\
 Ours       & 0.7710 & 0.2533 & \textbf{0.0003}  & \textbf{0.0159} \\
\bottomrule
\end{tabular}
\end{table}
\vspace{1mm}

\begin{figure}[htbp]
\def\lw{0.23\linewidth}
\def\hlw{0.065\lw} 
\renewcommand\tabcolsep{0.0pt}
\renewcommand{\arraystretch}{0}
\centering \small
\begin{tabular}{ccccc} 
\rotatebox{90}{\hspace{5.5mm}w/o DE} &
\includegraphics[width=\lw]{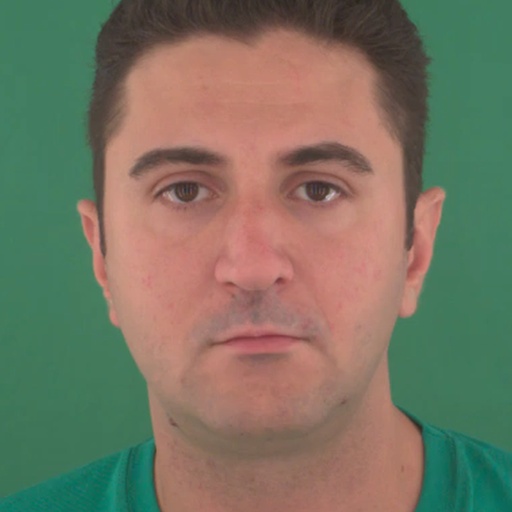} & 
\includegraphics[width=\lw]{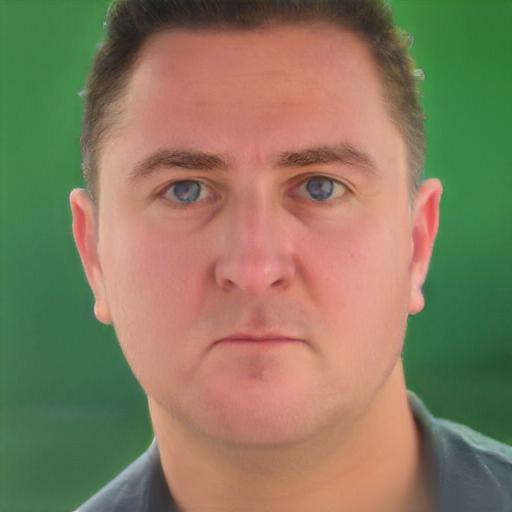} & 
\includegraphics[width=\lw]{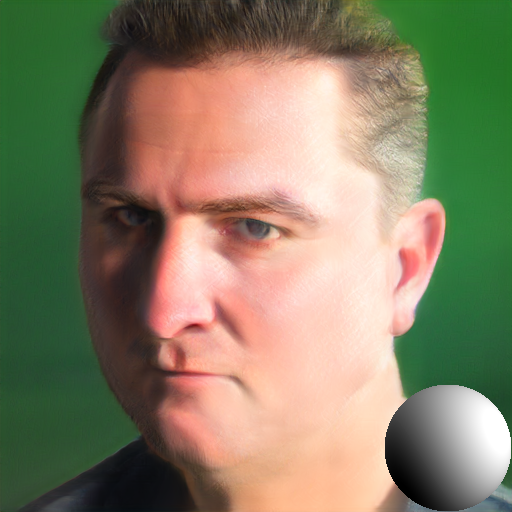} & 
\includegraphics[width=\lw]{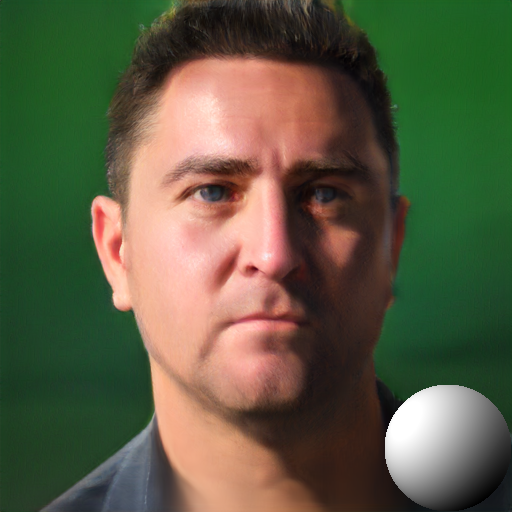}  \\
\rotatebox{90}{\hspace{5.5mm}ours} &
\includegraphics[width=\lw]{Figures/ablation/input.jpg} & 
\includegraphics[width=\lw]{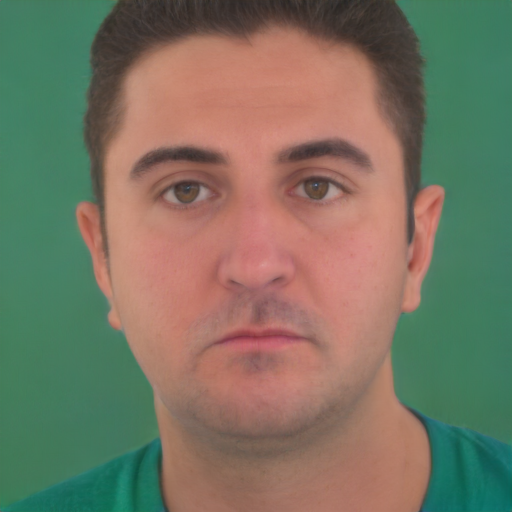} & 
\includegraphics[width=\lw]{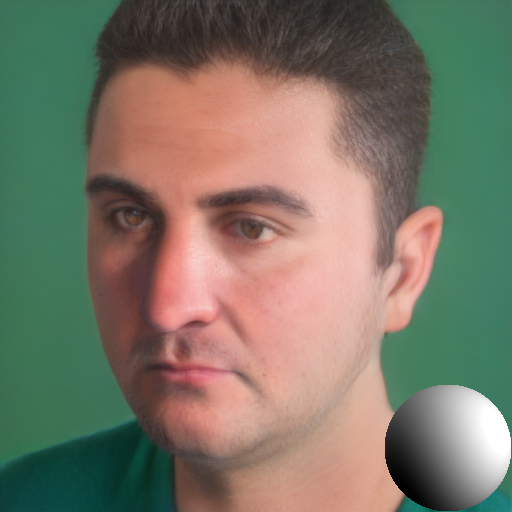} & 
\includegraphics[width=\lw]{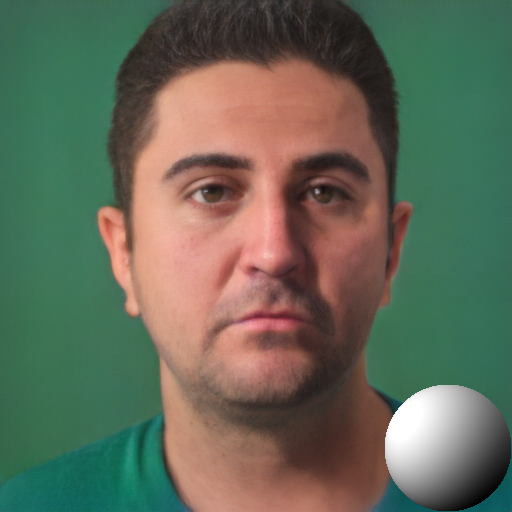}  \\
\vspace{0.3mm}
\\
 & Input & Reconstruction & Condition 1 & Condition 2
\end{tabular}
\vspace{-1mm}
\caption{Albation study comparing our model with and without the tri-plane encoders. The model without tri-plane encoders replaces our tri-plane encoders with an existing latent space encoder. This replacement results in images that bear much less resemblance to the input person, indicating a lower level of identity preservation.}
\label{fig:ablation study}
\end{figure}

\paragraph{Tri-plane Dual-Encoders Design.}
We remove the dual-encoders (DE) and use an existing latent code encoder from \cite{nerffacelighting} instead. While this alternative design does achieve real-time 3D-aware relighting, it comes at the cost of a substantial reduction in {reconstruction quality}%
, as visually depicted in Figure \ref{fig:ablation study}. 

\section{Conclusion, Limitations and Future Work}
\label{sec:Conclusion}

\paragraph{Conclusion.}
We introduced a real-time 3D-aware method for portrait video relighting and novel view synthesis. Our method can recover coherent and consistent geometry and relight the video under novel lighting conditions for a given facial video. 
Our method combines the benefits of a relightable generative model, \ie, disentanglement and controllability, to capture the intrinsic geometry and appearance of the face in a video and generate realistic and consistent videos under novel lighting conditions. We evaluated our method on portrait videos and showed its superiority over existing methods in terms of lighting accuracy and lighting stability. Our work opens up new possibilities for 3D-aware portrait video relighting and synthesis.

\paragraph{Limitations.}
{One of the limitations of our method is that it fails to model %
glares on the eyeglasses, {as shown in the rightmost column of \Fref{fig:input view relighting}. {Future enhancements could benefit from incorporating advanced reflection and refraction modeling techniques.} Furthermore, our method does not separate the motion information from the %
identity information, thus limiting its ability to perform video-driven animation.} 
{This challenge might be addressed through the integration of the latest advancements in talking head generation techniques.}

\paragraph{Future Work.}
We are interested in extending our method to handle more complex scenes, such as multiple faces, occlusions, and full-body relighting. We also intend to explore more applications of our method, such as face editing and animation. 

\vspace{-2mm}
\section*{Acknowledgement}
\vspace{-2mm}
This work was supported by National Natural Science Foundation of China (No. 62322210, No. 62102403, No. 62136001 and No. 62088102), Beijing Municipal Natural Science Foundation for Distinguished Young Scholars (No. JQ21013), and Beijing Municipal Science and Technology Commission (No. Z231100005923031). We thank Yu Li from the High Performance Computing Center at Beijing Jiaotong University for his support and guidance in parallel computing optimization. We also thank Yu-Ying Yeh for generously sharing data for comparison.

{
    \small
    \bibliographystyle{ieeenat_fullname}
    \bibliography{main}
}


\end{document}